\documentclass[letterpaper, 10 pt, conference]{ieeeconf}
\IEEEoverridecommandlockouts 
\usepackage{hyperref}
\usepackage{threeparttable}
\usepackage{algorithm}
\usepackage{cite}

\hypersetup{
    colorlinks=true,
    linkcolor=magenta,
    filecolor=magenta,      
    urlcolor=magenta,
}

\renewcommand{\thefigure}{\Roman{figure}}

\usepackage{graphics}           
\usepackage{times}              
\usepackage{amsmath}            
\usepackage{amssymb}            
\usepackage{graphicx}
\usepackage{algorithm}
\usepackage[noend]{algpseudocode}
\usepackage{booktabs}
\usepackage{color}
\definecolor{instructioncolor}{rgb}{.5,.5,.5}

\usepackage[font=small]{caption}

\def\eqref#1{Eq.~(\ref{#1})}

\makeatletter
\usepackage{xspace}
\DeclareRobustCommand\onedot{\futurelet\@let@token\@onedot}
\def\@onedot{\ifx\@let@token.\else.\null\fi\xspace}

\def\etal{{et al}\onedot}
\makeatother

\usepackage{array}
\newcolumntype{L}[1]{>{\raggedright\let\newline\\\arraybackslash\hspace{0pt}}m{#1}}
\newcolumntype{C}[1]{>{\centering\let\newline\\\arraybackslash\hspace{0pt}}m{#1}}
\newcolumntype{R}[1]{>{\raggedleft\let\newline\\\arraybackslash\hspace{0pt}}m{#1}}

\usepackage{bm}
\usepackage{multirow}
\usepackage{rotating}

\title{\LARGE \bf Spatiotemporal Decoupling for Efficient Vision-Based \\ Occupancy Forecasting}

\author{{Jingyi Xu$^{1 *}$, Xieyuanli Chen$^{2 *}$, Junyi Ma$^{1}$, Jiawei Huang$^{3}$, Jintao Xu$^{3}$, Yue Wang$^{4}$ and Ling Pei$^{1 \dag}$} 
  \thanks{$^{1}$Jingyi Xu, Junyi Ma, and Ling Pei are with the Shanghai Jiao Tong University. $^{2}$Xieyuanli Chen is with the National University of Defense Technology. $^{3}$Jiawei Huang, and Jintao Xu are with the HAOMO.AI. $^{4}$Yue Wang is with the Zhejiang University. $^{*}$Equal contribution}
  \thanks{$^\dag$Corresponding author: ling.pei@sjtu.edu.cn}
}

\begin{document}
\maketitle

\IEEEpeerreviewmaketitle
\thispagestyle{empty}
\pagestyle{empty}

\begin{abstract}

The task of occupancy forecasting (OCF) involves utilizing past and present perception data to predict future occupancy states of autonomous vehicle surrounding environments, which is critical for downstream tasks such as obstacle avoidance and path planning. Existing 3D OCF approaches struggle to predict plausible spatial details for movable objects and suffer from slow inference speeds due to neglecting the bias and uneven distribution of changing occupancy states in both space and time.
In this paper, we propose a novel spatiotemporal decoupling vision-based paradigm to explicitly tackle the bias and achieve both effective and efficient 3D OCF.
To tackle spatial bias in empty areas, we introduce a novel spatial representation that decouples the conventional dense 3D format into 2D bird’s-eye view (BEV) occupancy with corresponding height values, enabling 3D OCF derived only from 2D predictions thus enhancing efficiency.
To reduce temporal bias on static voxels, we design temporal decoupling to improve end-to-end OCF by temporally associating instances via predicted flows.
We develop an efficient multi-head network EfficientOCF to achieve 3D OCF with our devised spatiotemporally decoupled representation.
A new metric, conditional IoU (C-IoU), is also introduced to provide a robust 3D OCF performance assessment, especially in datasets with missing or incomplete annotations. The experimental results demonstrate that EfficientOCF surpasses existing baseline methods on accuracy and efficiency, achieving state-of-the-art performance with a fast inference time of 82.33\,ms with a single GPU. Our code will be released as open source.

\end{abstract}

\section{Introduction}
\label{sec:intro}

\begin{figure}
  \centering
  \captionsetup{aboveskip=2pt, belowskip=0pt}
  \includegraphics[width=1.0\linewidth]{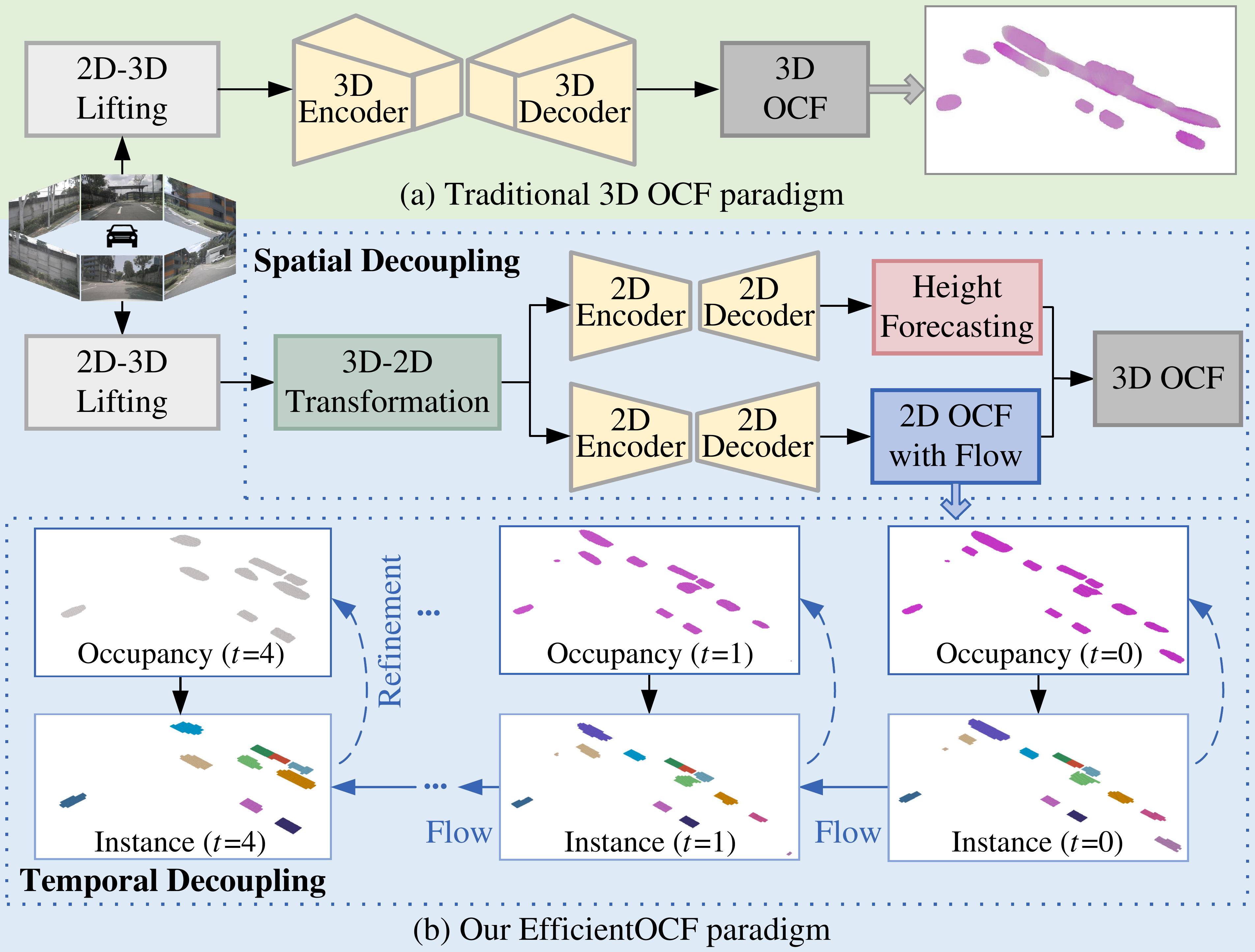}
  \caption{Unlike the traditional 3D vision-based OCF pipeline, our proposed paradigm implements spatiotemporal decoupling, thus achieving the efficiency of 2D OCF methods with the instance-aware capability of 3D OCF approaches.}
  \label{fig: motivation}
  \vspace{-0.6cm}
\end{figure}

Accurately estimating occupancy states in the surrounding environment based on visual inputs is crucial for autonomous mobile systems, as it enables the identification of general obstacles, including objects absent from semantic databases. Most existing vision-based occupancy estimation methods~\cite{cao2022monoscene, wang2023openoccupancy, li2023voxformer, zhang2023occformer, huang2023tri, hong2024univision, huang2024selfocc} rely on complex model architectures but often overlook temporal changes in occupancy states. To address this, occupancy forecasting (OCF) approaches~\cite{khurana2023point, hu2021fiery, akan2022stretchbev, zhang2022beverse, li2023powerbev, ma2024cam4docc} incorporate the prediction of surrounding object movements over time. These methods use past observations to forecast occupancy states in the present and near future, providing valuable information in advance for downstream tasks such as planning and control in autonomous driving.

Existing vision-based OCF methods predict future occupancy states for the entire 3D scene, as illustrated in Fig.~\ref{fig: motivation}(a). These methods process all voxel grids in each scene through end-to-end 3D feature encoding and decoding~\cite{ma2024cam4docc, liu2023lidar}. 
However, this indiscriminate dense representation often leads to inefficient and less accurate OCF due to two main biases: (1) a spatial bias toward empty voxels, which dominate the 3D space; (2) a temporal changing bias toward static objects, as only a small subset of movable objects change positions in near future. These biases not only result in high computational costs and wasteful resource use but also increase uncertainty in predicting movable object occupancy, leading to rapid error accumulation in forecasts.

To tackle these issues, we propose a novel spatiotemporal decoupling representation and an efficient network EfficientOCF for vision-based OCF. 
As shown in Fig.\,\ref{fig: motivation}(b), it decouples 3D OCF both spatially and temporally. We design a new spatial representation that decouples bird’s-eye view (BEV) occupancy with height values. By storing only heights for 2D occupied grids, it achieves 3D OCF while avoiding the traditional memory-intensive 3D occupancy representation~\cite{wang2023openoccupancy, wei2023surroundocc, pan2023uniocc, liu2024occtransformer}. 
For temporal changing forecasting, we decouple current instance segmentation from future occupancy estimation, leveraging instance movement understanding to enhance OCF rather than relying on end-to-end forecasting. 
To achieve these goals, we design EfficientOCF, a novel network with flow, segmentation, and height prediction heads. This network compresses 3D voxel features into 2D BEV features and employs lightweight 2D encoder-decoder structures to predict BEV occupancy. Simultaneously, it predicts 2D backward centripetal flow and height values for grids containing movable objects at each timestep. Flow vectors establish instance associations between current observations and future forecasts, refining OCF results at the instance level for enhanced accuracy and efficiency. The final 3D OCF is obtained by lifting the refined 2D OCF with height estimation.
We further introduce new metrics and conduct extensive experiments on the nuScenes \cite{caesar2020nuscenes}, nuScenes-Occupancy \cite{wang2023openoccupancy}, and Lyft-Level5 \cite{lyft2019} datasets to demonstrate that our EfficientOCF generates high-quality OCF results and achieves superior real-time performance compared to the state-of-the-art (SOTA).

In summary, the contributions of our work are threefold:

\begin{itemize}

\item We propose a spatiotemporal decoupling framework for effective and efficient vision-based 3D OCF. Spatial decoupling is designed with a representation that combines BEV occupancy with height values to avoid 3D voxel redundancy. Temporal decoupling is achieved via instance-aware associations, obtaining enhanced 3D OCF results.

\item We introduce EfficientOCF, a novel efficient vision-based 3D OCF network, that predicts the future occupancy states of movable objects using our proposed spatiotemporally decoupled representation. EfficientOCF employs lightweight 2D encoder-decoder structures rather than 3D counterparts and incorporates an adaptive dual pooling strategy for 3D-2D feature transformation, achieving SOTA 3D OCF performance and efficiency. 

\item We propose new evaluation metrics tailored for OCF on fine-grained general movable objects. These metrics address the shortcomings of traditional intersection over union (IoU) by minimizing penalties for false positives within the bounding box of movable objects, which enhances the rationality of assessments, especially in cases of missing or incomplete occupancy annotations. 

\end{itemize}

\section{Related Work}
\label{sec:related}

\textbf{3D occupancy prediction.}~3D occupancy prediction aims to estimate whether each voxel in the surrounding environment is currently occupied, representing general obstacles and enhancing the expressiveness of complex scenes.
MonoScene, proposed by Cao~\etal~\cite{cao2022monoscene}, first introduces a 3D semantic scene completion framework using only camera inputs, but it primarily focuses on front-view scenarios. To address this, TPVFormer by Huang~\etal~\cite{huang2023tri} proposes a tri-perspective view (TPV) representation to lift image features, achieving comparable performance with other LiDAR methods. Building on the TPV representation, they further propose SelfOcc~\cite{huang2024selfocc}, employing a self-supervised scheme to learn 3D occupancy states. 
Similar to TPVFormer, transformer modules~\cite{vaswani2017attention} are also utilized in Occ3D~\cite{tian2024occ3d}, OccFormer~\cite{zhang2023occformer}, and SurroundOcc~\cite{wei2023surroundocc}. Occ3D~\cite{tian2024occ3d} proposes CTF-Occ, a novel transformer-based course-to-fine occupancy network, achieving 3D occupancy predictions with high spatial resolution. OccFormer~\cite{zhang2023occformer} proposes a dual-path transformer network to fully integrate 2D global features and 3D local features, while SurroundOcc proposed by Wei~\etal~\cite{wei2023surroundocc} adopts spatial 2D-3D attention to lift images to the 3D volume space. 
Wang~\etal~\cite{wang2023openoccupancy} propose OpenOccupancy, the first surrounding semantic occupancy perception benchmark, and they introduce nuScenes-Occupancy dataset extending the large-scale nuScenes dataset~\cite{caesar2020nuscenes} with fine-grained occupancy annotations.
These approaches consider the complex challenge of inferring the spatial distribution of obstacles but ignore their temporal evolution. However, understanding the detailed and dynamic environment is crucial for planning and decision-making in autonomous driving, ensuring robust and reliable performance under various driving conditions.

\textbf{BEV occupancy forecasting.}~BEV representation has been widely used in autonomous mobile systems due to its lightweight 2D format~\cite{li2022bevformer, li2023bevdepth, liu2023bevfusion,liu2023multi, mahjourian2022occupancy}. 
To overcome the challenge of capturing the temporal evolution in 2D format, some occupancy perception methods are proposed to achieve occupancy forecasting and instance prediction in BEV.
FIERY, proposed by Hu~\etal~\cite{hu2021fiery}, is the first to use surround-view camera inputs to directly predict future instance segmentation and the motion of dynamic agents using BEV representation. 
Based on FIERY, StretchBEV by Akan~\etal~\cite{akan2022stretchbev} incorporates a stochastic temporal model to refine residual updates. Zhang~\etal~\cite{zhang2022beverse} propose BEVerse, which also uses BEV occupancy to jointly reason about multiple tasks, including object detection, semantic map construction, and motion prediction. 
Similarly, MotionNet by Wu~\etal~\cite{wu2020motionnet} encodes object category and motion information in each BEV grid to predict scene flow and detect objects. Compared to MotionNet, MP3 proposed by Casas~\etal~\cite{casas2021mp3} enhances occupancy flow parameterization with probabilistic modeling for multi-modal behavior.
In contrast, PowerBEV by Li~\etal~\cite{li2023powerbev} relies sorely on a semantic segmentation head alongside a backward centripetal flow head to achieve future instance prediction.
Despite their effectiveness, these BEV-based methods lack a comprehensive understanding of spatial structures along the z-axis, which is crucial for accurately capturing the full 3D states of obstacles.

\textbf{3D occupancy forecasting.}~Recently, researchers have increasingly explored end-to-end occupancy forecasting in 3D space. 
As depicted in the previous work~\cite{ma2024cam4docc}, occupancy states can be directly generated from point cloud prediction \cite{lu2021monet, luo2023pcpnet, pal2024atppnet, mersch2022self, fan2019pointrnn} by trivial voxelization, but cannot achieve acceptable occupancy forecasting accuracy. 
OccWorld~\cite{zheng2023occworld} and OccSora~\cite{wang2024occsora} use generation models to simultaneously predict the movement of the ego car and the evolution of the surrounding scenes. 
Ma~\etal~\cite{ma2024cam4docc} introduce Cam4DOcc, the first benchmark to standardize evaluation protocol of vision-only occupancy forecasting in 3D space, and propose OCFNet, an end-to-end network to forecast dense 3D occupancy states.
Compared to these 3D occupancy forecasting methods, in this work, we decouple spatiotemporal representation, where BEV occupancy with height values is utilized for spatial decoupling and instance-aware refinement in consecutive frames is designed for temporal decoupling. Based on the spatiotemporally decoupled representation, we propose a novel lightweight network EfficientOCF. Our experiments show that our proposed method achieves the SOTA performance with fast inference speed.

\begin{figure*}
  \centering
  \captionsetup{aboveskip=2pt, belowskip=0pt}
  \includegraphics[width=0.95\linewidth]{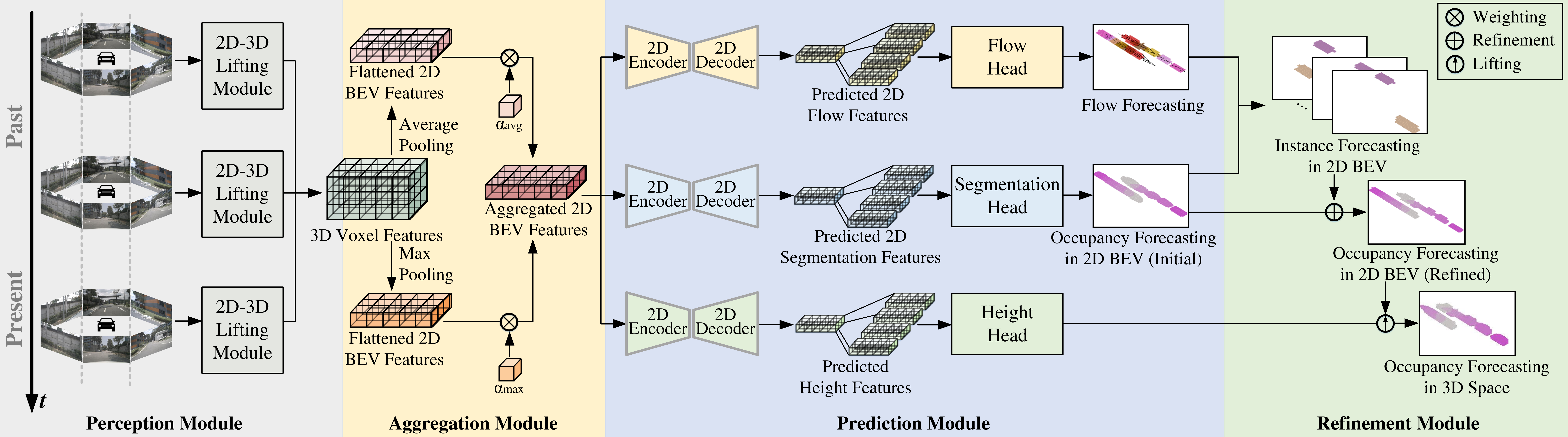}
  \caption{Architecture of our proposed EfficientOCF. The perception module (gray area) extracts 2D visual features from the input surrounding images and lifts them to 3D voxel features. Subsequently, the voxel features are further compressed into BEV features and aggregated by the aggregation module (yellow area). Next, the prediction module (blue area) with three branches sharing the same 2D structure, achieves initial OCF in 2D BEV and predicts backward centripetal flow and height values for BEV occupancy. The refinement module (green area) forecasts instances using estimated flow to refine initial 2D OCF results, which are ultimately lifted to 3D space.}
  \label{fig: network}
  \vspace{-0.4cm}
\end{figure*}

\section{Proposed Method}
\label{sec:EfficientOCF}

\subsection{Task Definition}
\label{sec:task}
\vspace{-0.1cm}

Given $N_\text{p}$ past and present surrounding visual images $\{\mathcal{I}|I_t\in\mathbb {R}^{N_\text{cam}\times 3\times H_\text{v}\times W_\text{v}}\}$, $t\in[-N_\text{p}, 0]$ as input, the vision-based OCF task aims to predict the occupancy states (occupied or free) of general movable objects in the present frame and $N_\text{f}$ future frames $\{\mathcal{O}^\text{3D}|O_t^\text{3D}\in\mathbb {R}^{1\times H\times W\times L}\}$, $t\in[0, N_\text{f}]$. $N_\text{cam}$ represents surrounding camera numbers. $H_\text{v}$, $W_\text{v}$ represent the height and width of input images, and $H$, $W$, $L$ represent the height, width, and length of the specific range defined in the present ($t=0$) coordinate system.

Compared to traditional prediction methods, our work first spatially decouples OCF in 2D BEV occupancy $\{\mathcal{O}^\text{2D}|O_t^\text{2D}\in\mathbb {R}^{1\times H\times W}\}$, $t\in[0, N_\text{f}]$ and corresponding height prediction $\{\mathcal{O}^\text{height}|O_t^\text{height}\in\mathbb {R}^{1\times H\times W}\}$, $t\in[0, N_\text{f}]$. It predicts the future height values of each voxel column and then lifts $\mathcal{O}^\text{2D}$ to $\mathcal{O}^\text{3D}$.
Unlike static objects that remain unchanged over short periods, general movable objects exhibit more dynamic motion characteristics. Therefore, following~\cite{ma2024cam4docc}, we focus on forecasting the occupancy states of general movable objects, e.g., bicycles, buses, cars, etc., in the nuScenes dataset~\cite{caesar2020nuscenes}. 
This emphasis on movable objects enhances the relevance and applicability of the forecasting task to real-world autonomous driving scenarios, where accurately predicting the motion of these objects is vital for both safety and efficiency. 
While existing 3D OCF methods~\cite{ma2024cam4docc} forecast movable objects end-to-end, we propose temporal decoupling to propagate current instance segmentation through flow association and achieve refined 3D OCF. The details of our approach are as follows.

\subsection{Network Architecture}
\label{sec:network}
\vspace{-0.1cm}

The pipeline of our EfficientOCF is shown in Fig.~\ref{fig: network}, which consists of four main components: a perception module, an aggregation module, a prediction module, and a refinement module.
The perception module extracts features from the input surrounding images and lifts them from 2D to 3D space following Lift-Splat-Shoot~\cite{philion2020lift} to obtain 3D spatial voxel features. The aggregation module then aggregates voxel features from sequential frames and transforms them into 2D BEV features, which are fed into the following prediction module. Next, the prediction module, composed of multiple 2D branches, decodes the aggregated 2D BEV features to initial 2D OCF, backward centripetal flow, and height values respectively. Then, the refinement module propagates instance segmentation into each future timestep, which is further used to refine initial OCF results in 2D BEV. Ultimately, we use height values predicted by the prediction module to lift the refined 2D occupancy to 3D space. 

\textbf{Perception module.}~To extract 3D voxel features from the input surrounding images, we adopt the approach used in previous studies~\cite{li2023powerbev, ma2024cam4docc}, applying a shared ResNet \cite{he2016deep} backbone to derive 2D features from both past and present images. These features are represented as $\{\mathcal{F}^\text{2D}|F_t^\text{2D}\in\mathbb {R}^{c_\text{v}\times h_\text{v}\times w_\text{v}}\}$, where $t\in[-N_\text{p}, 0]$, $c_\text{v}$ represents the 2D feature channel, and $h_\text{v}\times w_\text{v}$ denotes the 2D spatial resolution.
These 2D features are then lifted and integrated into 3D voxel features $\{\mathcal{F}^\text{3D}|F_t^\text{3D}\in\mathbb {R}^{c\times h\times w\times l}\}$ using a 2D-3D lifting module \cite{philion2020lift}, where $c$ is the 3D feature channel, and $h\times w\times l$ represents the 3D spatial resolution.

\textbf{Aggregation module.}~All the obtained 3D voxel features are projected into the present ($t\!=\!0$) coordinate system exploiting 6-DOF ego-car poses. Here we propose an adaptive dual pooling strategy to further transform 3D voxel features into 2D BEV features.
We first focus on the holistic information of each voxel column and employ average pooling to generate 2D features $\{\mathcal{F}^\text{avg}|F_t^\text{avg}\in\mathbb {R}^{c\times h\times w}\}$, where $t\in[-N_\text{p}, 0]$. Additionally, we use max pooling to generate the concurrent 2D features $\{\mathcal{F}^\text{max}|F_t^\text{max}\in\mathbb {R}^{c\times h\times w}\}$, capturing the voxel grid with prominent occupancy features implicitly corresponding to height values. 
Learnable weights, $\alpha_\text{avg}$ and $\alpha_\text{max}$, are designed to adaptively adjust the contribution of these two types of 2D BEV features $\mathcal{F}^\text{avg}$ and $\mathcal{F}^\text{max}$.
Finally, we aggregate features from both $N_\text{p}$ past and present inputs to obtain the aggregated 2D BEV features $F^\text{agg}\in\mathbb {R}^{(N_\text{p}+1)\times c\times h\times w}$.

\begin{figure}
  \centering
  \captionsetup{aboveskip=1pt, belowskip=0pt}
  \includegraphics[width=0.9\linewidth]{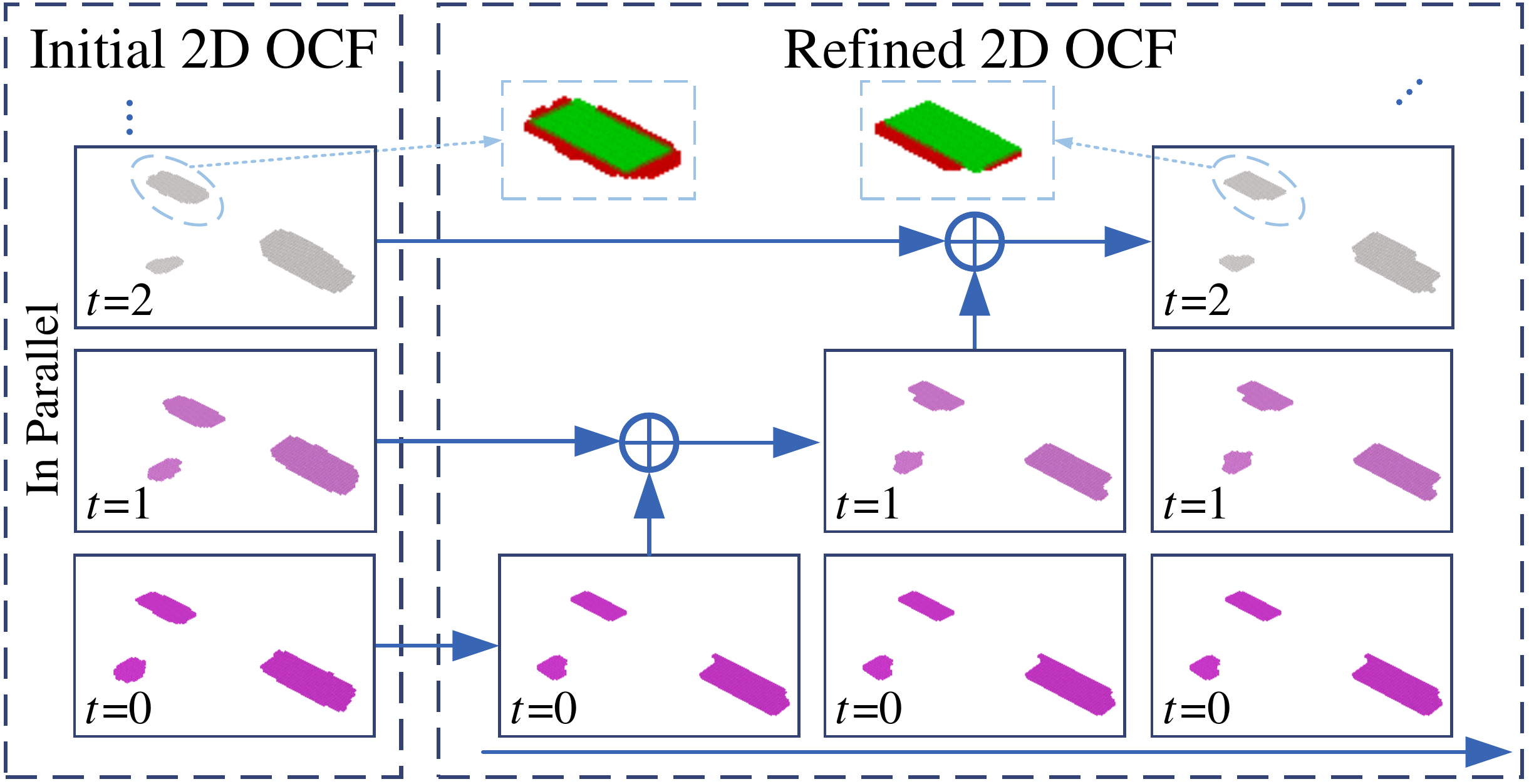}
  \caption{The process of refining initial OCF results following temporal decoupling. We exploit step-wise association along the time axis to improve forecasting quality.}
  \label{fig: decoupling}
  \vspace{-0.6cm}
\end{figure}

\textbf{Prediction module.}~The proposed prediction module consists of three 2D convolution-based branches for forecasting initial 2D occupancy, height values, and backward centripetal flow vectors respectively. 
The aggregated 2D BEV features are fed into 2D encoder-decoder structures, and then processed by three predictive heads. 
The segmentation head estimates initial present and future occupancy states $\{\mathcal{O}^\text{2D}|O_t^\text{2D}\in\mathbb {R}^{1\times H\times W}\}$, $t\in[0, N_\text{f}]$ in 2D space. Concurrently, the height head determines the height values of 2D voxel grids, denoted as $\{\mathcal{O}^\text{height}|O_t^\text{height}\in\mathbb {R}^{1\times H\times W}\}$ for $t\in[0, N_\text{f}]$, based on our proposed representation with spatial decoupling.
In addition, the flow head generates 2D backward centripetal flow \cite{li2023powerbev} for each voxel grid between consecutive frames, denoted as $\{\mathcal{O}^\text{flow}|O_t^\text{flow}\in\mathbb {R}^{2\times H\times W}\}$, $t\in[0, N_\text{f}]$, which points from a voxel at time $t$ to its corresponding potential 2D instance center at $t-1$. 

\begin{figure}
  \centering
  \captionsetup{aboveskip=2pt, belowskip=0pt}
  \includegraphics[width=0.9\linewidth]{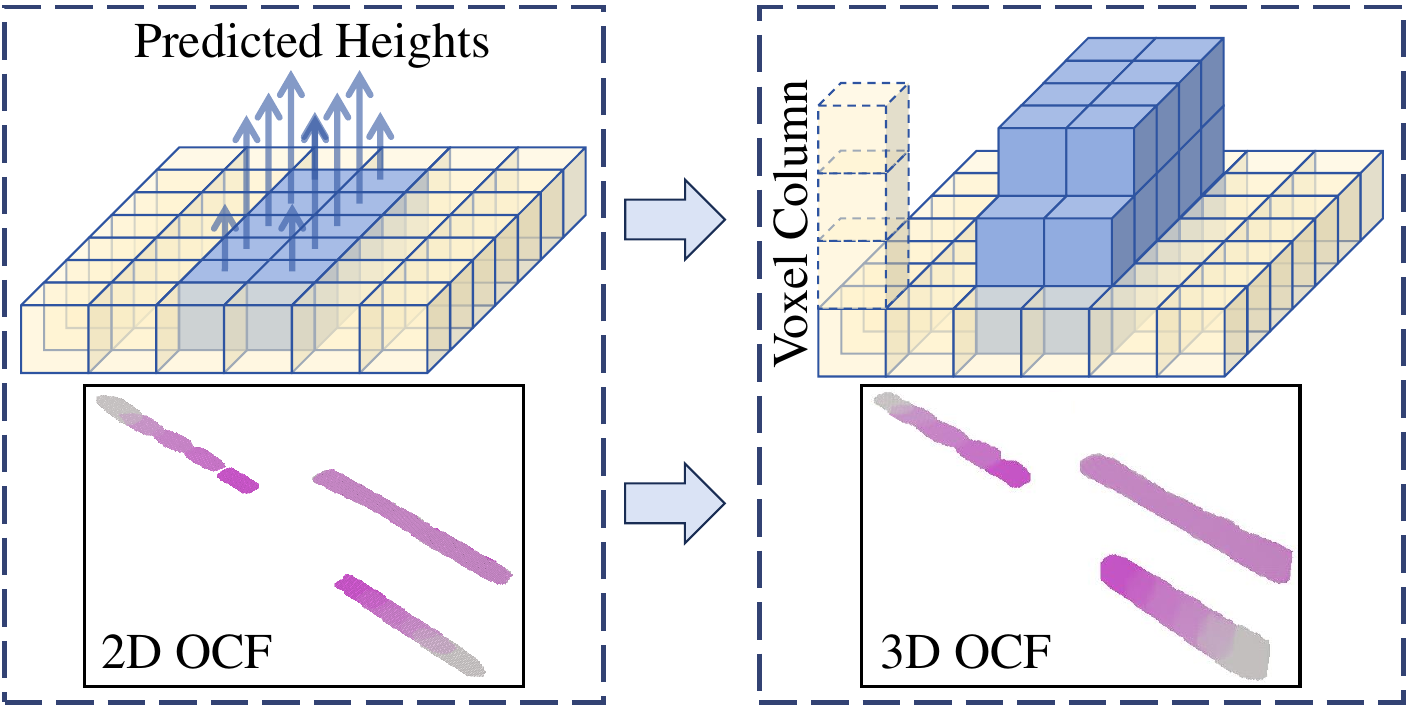}
  \caption{EfficientOCF achieves 3D OCF by assigning 2D grids with refined occupancy states and heights changing with time.}
  \label{fig: lifting}
  \vspace{-0.4cm}
\end{figure}

\textbf{Refinement module.}~
We find that existing OCF methods~\cite{li2023powerbev,ma2024cam4docc} suffer from shape divergence over time due to insufficient instance consistency across temporal frames.
To address this, we propose decoupling current instance segmentation from future occupancy forecasts and refining 3D OCF through instance association, rather than learning 3D OCF end-to-end. The motivation is that instance segmentation based on current observations is often more accurate than end-to-end forecasts for future unobserved frames. By separating segmentation from occupancy forecasting, we simplify the learning process, making it easier compared to end-to-end 3D OCF. This allows us to leverage more accurate instance shapes to improve 3D OCF results.
To this end, we design a refinement module that first performs instance segmentation in 2D BEV and then uses associated instances to mask inaccurate occupancy predictions for movable objects in the initial 2D OCF results, which are then lifted by height information to obtain refined 3D OCF. 
Specifically, we extract the centers of instances by non-maximum suppression (NMS) at $t\!=\!-1$ and associate pixel-wise instance ID between adjacent frames iteratively over time $t\in[0, N_\text{f}]$ using the predicted 2D backward centripetal flow. Here we denote the instance forecasting results as $\{\mathcal{M}^\text{2D}|M_t^\text{2D}\in\mathbb {R}^{1\times H\times W}\}$, $t\in[0, N_\text{f}]$. For each timestep $t$, we convert $M_t^\text{2D}$ to the instance mask $\bar{M}_t^\text{2D}$ by $\mathtt{CLIP}(M_t^\text{2D})$, which restricts instance IDs exceeding 1 to 1. Ultimately, we generate the final refined 2D OCF results $\bar{\mathcal{O}}^\text{2D}$ by $O_t^\text{2D} \cdot \bar{M}_t^\text{2D}$ at each timestep, as Fig.~\ref{fig: decoupling} shows.
We therefore achieve temporal decoupling by splitting the end-to-end OCF into step-wise instance associations along the time axis.
By assigning each 2D voxel grid with corresponding refined occupancy states and height values over time, we achieve lifting 2D OCF $\bar{\mathcal{O}}^\text{2D}$ to 3D OCF $\{\bar{\mathcal{O}}^\text{3D}|\bar{O}_t^\text{3D}\in\mathbb {R}^{1\times H\times W\times L}\}$, $t\in[0, N_\text{f}]$, as shown in Fig.~\ref{fig: lifting}.

\subsection{Training and Inference}
\label{sec:training}
\vspace{-0.1cm}

Our approach is trained and evaluated on three publicly available datasets, including the nuScenes \cite{caesar2020nuscenes}, nuScenes-Occupancy \cite{wang2023openoccupancy}, and Lyft-Level5 \cite{lyft2019} datasets.

\textbf{Data generation.}~We develop a novel data generation method based on three publicly available datasets. 
Initially, we split these datasets into sequences with a time length of $N = N_\text{p} + N_\text{f} + 1$. Sequential semantic and instance annotations of general movable objects are then extracted for each sequence. 
Based on the extracted semantic information, we set the voxel grids corresponding to the general movable objects to $1$ and the other positions to $0$. We thereby obtain the bounding-box-aware 3D occupancy states $\{\hat{\mathcal{O}}^\text{bb}|\hat{O}_t^\text{bb}\in\mathbb {R}^{1\times H\times W\times L}\}$ (where $1$ indicates occupied and $0$ indicates free), which corresponds to the inflated annotations mentioned in \cite{ma2024cam4docc}. In this case, we assign the heights of ground-truth bounding boxes to the voxel grids within them.
Subsequently, we compress 3D voxel annotations into 2D BEV counterparts $\{\hat{\mathcal{O}}^\text{bev}|\hat{O}_t^\text{bev}\in\mathbb {R}^{1\times H\times W}\}$. This is achieved under the principle that if any voxel in a vertical voxel column has a value of $1$, the corresponding position in the BEV is set to $1$. We extract ground-truth height values of each voxel column $\{\hat{\mathcal{O}}^\text{height}|\hat{O}_t^\text{height}\in\mathbb {R}^{1\times H\times W}\}$ by selecting the highest grid with the semantics of general movable objects.
We then conduct fine-grained 3D occupancy states $\{\hat{\mathcal{O}}^\text{fg}|\hat{O}_t^\text{fg}\in\mathbb {R}^{1\times H\times W\times L}\}$ by assigning height values to 2D BEV occupancy annotations.

\textbf{Loss functions.}~We employ the cross-entropy loss as the OCF loss $\mathcal{L}_{\text{occ}}$, and use the smooth L1 loss for both the height prediction loss $\mathcal{L}_{\text{height}}$ and the flow prediction loss $\mathcal{L}_{\text{flow}}$.
The total loss used to train our EfficientOCF is the weighted sum of these three losses, computed as follows:
\begin{align}
\small
\begin{split}
    \mathcal{L}_\text{all} & = \frac{1}{N_\text{f}+1}\sum_{t=0}^{N_\text{f}}(\lambda_\text{1} \mathcal{L}_{\text{occ}}(\hat{{O}}_t^\text{bev}, \bar{{O}}_t^\text{2D}) \\
    & + \lambda_\text{2} \mathcal{L}_{\text{height}}(\hat{{O}}_t^\text{height}, {O}_t^\text{height}) + \lambda_\text{3} \mathcal{L}_{\text{flow}}(\hat{{O}}_t^\text{flow}, {O}_t^\text{flow})),
\end{split}
\label{eq:loss}
\end{align}
where $\lambda_\text{1}$, $\lambda_\text{2}$, and $\lambda_\text{3}$ are the weighting parameters to balance the supervision for different heads.

\begin{figure}
  \centering
  \captionsetup{aboveskip=2pt, belowskip=0pt}
  \includegraphics[width=0.8\linewidth]{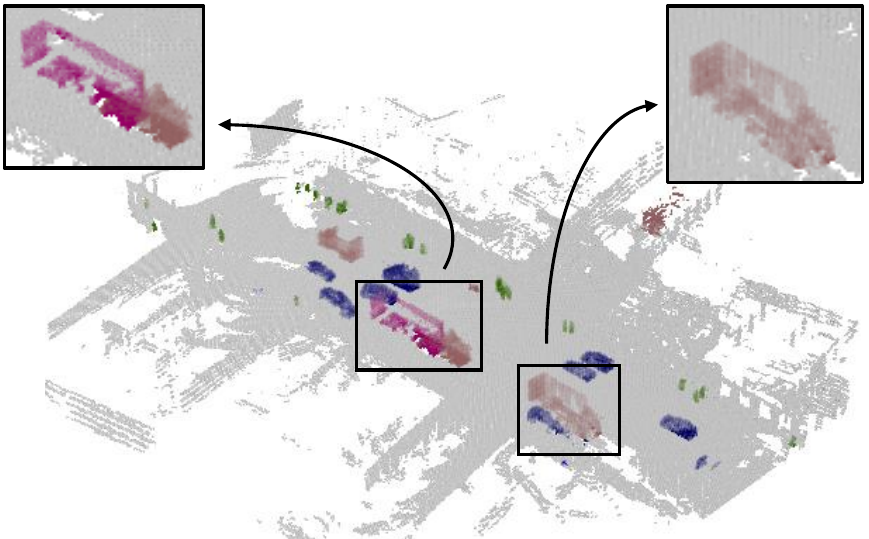}
  \caption{Examples of incomplete 3D occupancy labels due to LiDAR sparsity in nuScenes-Occupancy \cite{wang2023openoccupancy}, which cannot comprehensively assess occupancy forecasting methods.}
  \label{fig: data}
  \vspace{-0.4cm}
\end{figure}

\textbf{Evaluation metrics.}~We employ intersection-over-union (IoU), and conditional IoU (C-IoU) to evaluate OCF performance. We also use video panoptic quality (VPQ)~\cite{kim2020video} to assess the performance of 3D instance forecasting.
We utilize $\text{IoU}_\text{c}$, $\text{IoU}_\text{f}$, and $\tilde{\text{IoU}}_\text{f}$ to separately evaluate the occupancy estimation at the present moment ($t = 0$), the forecasting results for future timesteps ($t\in[1, N_\text{f}]$), and the occupancy estimation over the entire time horizon ($t\in[0, N_\text{f}]$). These evaluation metrics are computed by
\begin{align}
\small
    \text{IoU} = \frac{1}{N_\text{all}}\sum_{t=t_\text{s}}^{t_\text{e}}\frac{\sum_{}\hat{{O}}_t^\text{occ}\cdot\bar{{O}}_t^\text{occ}}{\sum_{}\hat{{O}}_t^\text{occ}+\bar{{O}}_t^\text{occ}-\hat{{O}}_t^\text{occ}\cdot\bar{{O}}_t^\text{occ}}. 
\label{eq:iou_f}
\end{align}
We set $\hat{{O}}_t^\text{occ}\text{/}\bar{{O}}_t^\text{occ}$ as $\hat{{O}}_t^\text{bev}\text{/}\bar{{O}}_t^\text{2D}$ to access 2D OCF performance.
To evaluate the performance of forecasting 3D inflated and fine-grained occupancy respectively, we calculate IoU by setting $\hat{{O}}_t^\text{occ}\text{/}\bar{{O}}_t^\text{occ}$ as $\hat{{O}}_t^\text{bb}\text{/}\bar{{O}}_t^\text{3D}$ and $\hat{{O}}_t^\text{fg}\text{/}\bar{{O}}_t^\text{3D}$. 
We set $t_\text{s}=t_\text{e}=0$ and $N_\text{all}=1$ to calculate  $\text{IoU}_\text{c}$, $t_\text{s}=1$, $t_\text{e}=N_\text{f}$, and $N_\text{all}=N_\text{f}$ for $\text{IoU}_\text{f}$, and $t_\text{s}=0$, $t_\text{e}=N_\text{f}$, and $N_\text{all}=N_\text{f}+1$ for $\tilde{\text{IoU}}$.

\begin{table*}[t]
\scriptsize
\setlength{\tabcolsep}{3.5pt}
\center
\captionsetup{aboveskip=2pt, belowskip=0pt}
\renewcommand\arraystretch{0.9}
\caption{Comparison of performance on 3D occupancy forecasting}
\begin{tabular}{l|ccc|ccc|ccc|ccc|cccccc}
\toprule
\multicolumn{1}{l|}{\multirow{3}{*}{Approach}}   & \multicolumn{6}{c|}{nuScenes} & \multicolumn{6}{c|}{Lyft-Level5} & \multicolumn{6}{c}{nuScenes \& nuScenes-Occupancy}\\ \cmidrule{2-19}  
\multicolumn{1}{c|}{}  & \multicolumn{3}{c|}{2D} & \multicolumn{3}{c|}{3D}   & \multicolumn{3}{c|}{2D}  & \multicolumn{3}{c|}{3D} & \multicolumn{6}{c}{3D} \\ \cmidrule{2-19} 
\multicolumn{1}{c|}{}                                                                               & $\text{IoU}_\text{c}$   & $\text{IoU}_\text{f}$    & $\tilde{\text{IoU}}$   & $\text{IoU}_\text{c}$    & $\text{IoU}_\text{f}$  & $\tilde{\text{IoU}}$    & $\text{IoU}_\text{c}$    & $\text{IoU}_\text{f}$   & $\tilde{\text{IoU}}$   & $\text{IoU}_\text{c}$    & $\text{IoU}_\text{f}$  & $\tilde{\text{IoU}}$  & $\text{IoU}_\text{c}$    & $\text{IoU}_\text{f}$  & $\tilde{\text{IoU}}$ & $\text{C-IoU}_\text{c}$    & $\text{C-IoU}_\text{f}$    & $\tilde{\text{C-IoU}_\text{f}}$\\ \cmidrule{1-19}
PowerBEV    & 33.62    & 30.07    & 30.77    & 28.34    & 25.47    & 26.05    & 45.78    & 43.44    & 43.91    & 36.15    & 34.18    & 34.58    & -    & -    & -    & -    & -   & -        \\
OpenOccupancy    & 28.77    & 25.93    & 26.50    & 24.26    & 22.69   & 23.15    & 38.16    & 35.11    & 35.79     & 32.95    & 29.73   & 30.40    & 14.99    & 13.06    & 13.47    & 27.11    & 24.68   & 25.17        \\ 
OccFormer    & 38.98    & 31.05    & 32.76    & 34.75    & 28.08    & 29.56    & 47.00    & 32.10    & 35.02    & 41.68    & 28.55   & 31.00    & 15.80    & 11.30    & 12.18    & 44.56    & 35.98   & 37.67        \\
OCFNet$^{-}$    & 30.34    & 26.27    & 27.47    & 27.32    & 24.40    & 24.98    & 36.02    & 33.12    & 33.69    & 31.04    & 28.92   & 29.36    & 12.66   & 10.65    & 11.05    & 29.64    & 27.47   & 27.91        \\ 
OCFNet    & 35.50   & 32.02    & 32.71    & 33.24    & 29.36    & 29.94    & 45.74    & 43.67    & 44.08    & 39.71    & 38.14   & 38.46    & 14.73    & 12.57    & 13.00    & 34.83    & 33.14   & 33.49        \\ \cmidrule{1-19}
EfficientOCF$^{-}$(ours)    & 31.69    & 29.00    & 29.56    & 30.24    & 28.17    & 28.60    & 36.39    & 33.24    & 33.83    & 33.95    & 31.12   & 31.70    & 15.98    & 14.05    & 14.46    & 35.89    & 34.20   & 34.56         \\
EfficientOCF (ours)    & \textbf{39.93}   & \textbf{36.15}  & \textbf{36.93}    & \textbf{35.60}    & \textbf{32.73}  & \textbf{33.32}      & \textbf{47.42}    & \textbf{44.96}    & \textbf{45.46}    & \textbf{44.27}    & \textbf{41.90}   & \textbf{42.39}     & \textbf{21.28}  & \textbf{19.02}  & \textbf{19.50}  & \textbf{47.53}  & \textbf{45.57}  & \textbf{45.99}     \\ \bottomrule
\end{tabular}
\label{tab: comparison}
\vspace{-0.4cm}
\end{table*}

While conventional IoU metrics can evaluate forecasting results within bounding boxes or fine-grained voxels \cite{ma2024cam4docc}, they struggle with scenarios involving incomplete 3D annotations due to LiDAR data sparsity, as illustrated in Fig.~\ref{fig: data}. The nuScenes-Occupancy dataset \cite{wang2023openoccupancy} extends the original nuScenes dataset \cite{caesar2020nuscenes} with dense semantic occupancy annotation using the proposed augmenting and purifying (AAP) pipeline. However, during the AAP's automatic labeling process, some annotation omissions are inevitable, resulting in missed 3D occupancy labels due to factors such as occlusion or sparse LiDAR observations, and even introducing pseudo occupancy labels with self-training \cite{xie2020self}. Additionally, during the manual purifying phase, while noise removal is effective, high-precision leak detection remains challenging. 
Therefore, we propose a new evaluation metric for 3D occupancy forecasting, conditional IoU ($\text{C-IoU}$):
\begin{align}
\small
     \text{C-IoU} = \frac{1}{N_\text{all}}\sum_{t=t_\text{s}}^{t_\text{e}}\frac{|TP_t|+|\widetilde{FP}_t|}{|TP_t|+|FN_t|+(|FP_t|-|\widetilde{FP}_t|))},
\label{eq:iou_f*}
\end{align}
where $TP_t$, $FN_t$, and $FP_t$ represent true positives, false negatives, and false positives at timestep $t$, computed in the 3D fine-grained format. $\widetilde{FP}_t$ represents partial false positives of $FP_t$, which lies in the corresponding annotated bounding boxes. Note that we only use this metric for nuScenes and nuScenes-Occupancy datasets because Lyft-Level5 lacks fine-grained voxel-wise annotations.
$\text{C-IoU}$ comprehensively considers predictions within bonding boxes and fine-grained occupancy predictions, accounting for the discrepancies between the nuScenes and nuScenes-Occupancy datasets to avoid penalizing false positives in bounding boxes unfairly. This mitigates the effect of missing and incomplete occupancy annotations within bounding boxes of movable objects.
Additionally, we use VPQ~\cite{kim2020video} to evaluate 3D instance forecasting lifted from 2D counterparts by height values, which is calculated by
\begin{align}
\small
     \text{VPQ}(\hat{\mathcal{O}}^\text{3D}, {\bar{\mathcal{O}}}^\text{3D}) = \sum_{t=0}^{N_\text{f}}\frac{\sum_{TP_t}\text{IoU}_t}{|TP_t|+\frac{1}{2}|FP_t|+\frac{1}{2}|FN_t|}.
\label{eq:vpq}
\end{align}
We set $\hat{\mathcal{O}}^\text{3D}$ as $\hat{\mathcal{O}}^\text{bb}$ and $\hat{\mathcal{O}}^\text{fg}$ for inflated and fine-grained formats, denoted as $\text{VPQ}^\text{bb}$ and $\text{VPQ}^\text{fg}$ respectively.

\begin{table*}[t]
\scriptsize
\setlength{\tabcolsep}{1.4pt}
\center
\captionsetup{aboveskip=2pt, belowskip=0pt}
\renewcommand\arraystretch{1}
\caption{Comparison of performance on 3D OCF in different future time horizons}
\begin{tabular}{l|cccc|cccc|cccc|cccc|cccc|cccc}
\toprule
\multicolumn{1}{l|}{\multirow{3}{*}{Approach}}   & \multicolumn{8}{c|}{nuScenes} & \multicolumn{8}{c|}{Lyft-Level5} & \multicolumn{8}{c}{nuScenes \& nuScenes-Occupancy}\\ \cmidrule{2-25}  
\multicolumn{1}{c|}{}  & \multicolumn{4}{c|}{$\text{IoU}_\text{f}\,(\text{2D})$} & \multicolumn{4}{c|}{$\text{IoU}_\text{f}\,(\text{3D})$}   & \multicolumn{4}{c|}{$\text{IoU}_\text{f}\,(\text{2D})$}  & \multicolumn{4}{c|}{$\text{IoU}_\text{f}\,(\text{3D})$} & \multicolumn{4}{c|}{$\text{IoU}_\text{f}\,(\text{3D})$}  & \multicolumn{4}{c}{$\text{C-IoU}_\text{f}\,(\text{3D})$} \\ \cmidrule{2-25} 
\multicolumn{1}{c|}{} & 0.5s  & 1.0s  & 1.5s  & 2.0s  & 0.5s  & 1.0s  & 1.5s  & 2.0s  & 0.2s  & 0.4s  & 0.6s  & 0.8s 
 & 0.2s  & 0.4s  & 0.6s  & 0.8s  & 0.5s  & 1.0s  & 1.5s  & 2.0s  & 0.5s  & 1.0s  & 1.5s  & 2.0s  \\ \cmidrule{1-25}
PowerBEV    & 32.13    & 31.31    & 30.61    & 30.07    & 27.17    & 26.49    & 25.92    & 25.47    & 45.04    & 44.64    & 44.05    & 43.44    & 35.52    & 35.18    & 34.68    & 34.18    & -    & -    & -    & -   & -   & -    & -    & -        \\
OpenOccupancy   & 27.89    & 27.11    & 26.46    & 25.93    & 24.20    & 23.79    & 23.26    & 22.69    & 37.24    & 36.62    & 35.86    & 35.11    & 32.03    & 31.19   & 30.45    & 29.73    & 14.08    & 13.71    & 13.36    & 13.06   & 26.62   & 25.83    & 25.16    & 24.68       \\ 
OccFormer    & 34.35    & 32.52    & 31.60    & 31.05    & 31.09    & 29.49    & 28.62    & 28.08    & 39.77    & 35.67    & 33.48    & 32.10    & 34.92    & 31.56   & 29.74    & 28.55    & 12.85    & 12.07    & 11.61    & 11.30   & 39.35   & 37.52    & 36.56    & 35.98       \\
OCFNet$^{-}$    & 29.11    & 28.11    & 27.38    & 26.77    & 26.31    & 25.51    & 24.90    & 24.40    & 35.05    & 34.46    & 33.80    & 33.12    & 30.39    & 29.97     & 29.45    & 28.92    & 11.65    & 11.25    & 10.94    & 10.65   & 29.09   & 28.39    & 27.88    & 27.47       \\ 
OCFNet    & 34.45    & 33.65    & 32.84    & 32.02    & 31.40    & 30.73    & 30.04    & 29.36    & 45.18    & 44.77    & 44.27    & 43.67    & 39.40    & 39.02    & 38.59    & 38.14    & 13.64    & 13.34    & 12.97    & 12.57   & 34.64   & 34.17    & 33.66    & 33.14       \\ \cmidrule{1-25}
EfficientOCF$^{-}$    & 30.90    & 30.16    & 29.52    & 29.00    & 29.74    & 29.13    & 28.60    & 28.17    & 35.35    & 34.64    & 33.93    & 33.24    & 32.88    & 32.43   & 31.78    & 31.12    & 15.07    & 14.70    & 14.35    & 14.05   & 35.86     & 35.16    & 34.61    & 34.20        \\
EfficientOCF    & \textbf{38.80}   & \textbf{37.85}  & \textbf{36.98}    & \textbf{36.15}    & \textbf{34.81}  & \textbf{34.07}  & \textbf{33.37}  & \textbf{32.73}  & \textbf{46.70}    & \textbf{46.20}  & \textbf{45.59}  & \textbf{44.96}  & \textbf{43.54}  & \textbf{43.07}   & \textbf{42.49}     & \textbf{41.90}  & \textbf{20.21}  & \textbf{19.86}  & \textbf{19.46}  & \textbf{19.02}  & \textbf{47.63}   & \textbf{46.91}    & \textbf{46.24}    & \textbf{45.57}      \\ \bottomrule
\end{tabular}
\label{tab: time}
\vspace{-0.2cm} 
\end{table*}

\begin{figure*}
  \centering
  \captionsetup{aboveskip=2pt, belowskip=0pt}
  \includegraphics[width=0.95\linewidth]{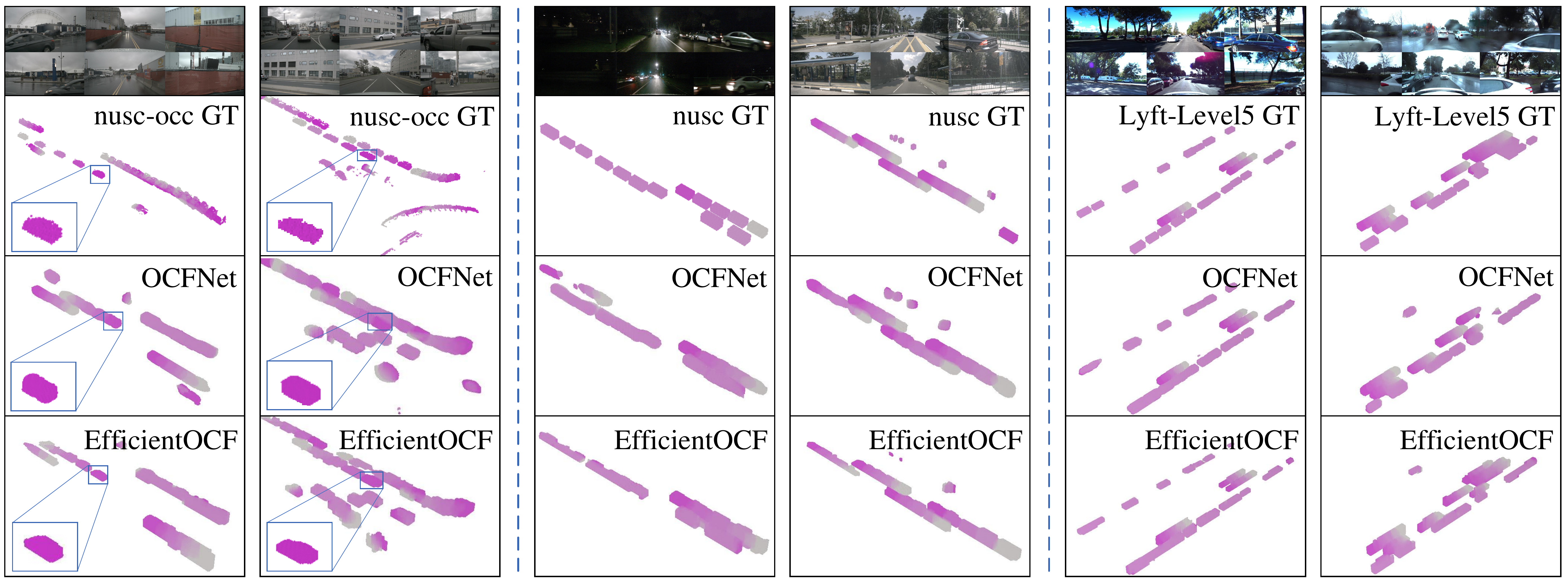}
  \caption{Visualization of OCF results of EfficientOCF and OCFNet, and ground-truth from nuScenes (nusc), nuScenes-Occupancy (nusc-occ), and Lyft-Level5 datasets. The occupancy forecasting results and ground-truth labels from timesteps 0 to $N_\text{f}$ are assigned colors from dark to light.}
  \label{fig: visualization}
  \vspace{-0.5cm}
\end{figure*}


\section{Experiments}
\label{sec:experiments}

\subsection{Experimental Setups}
\label{sec:setups}
\vspace{-0.1cm}

\textbf{Dataset details.}~We evaluate our proposed EfficientOCF alongside the SOTA baselines using the nuScenes \cite{caesar2020nuscenes}, nuScenes-Occupancy \cite{wang2023openoccupancy} and Lyft-Level5 \cite{lyft2019} datasets. Following existing works \cite{li2023powerbev, wang2023openoccupancy, ma2024cam4docc}, we split the nuScenes and nuScenes-Occupancy datasets into training and test sets with 700 and 150 scenes respectively, and the Lyft-Level5 dataset into 130 training scenes and 50 test scenes.
Each sequence has a time length $N$ of $7$ frames ($N_\text{p}=2$, $N_\text{f}=4$), meaning that we use $2$ past frames and $1$ present observation to estimate the present occupancy states and forecast the counterparts of the next $4$ frames.
The nuScenes and nuScenes-Occupancy datasets are labeled at a frequency of $2$$\,$Hz, and the Lyft-Level5 dataset at $5$$\,$Hz, resulting in different forecasting time durations across these datasets.

\textbf{Implementation details.}~Following \cite{ma2024cam4docc}, we set the evaluation range to [$-51.2\,$m, $51.2\,$m] for both the x-axis and y-axis, and [$-5\,$m, $3\,$m] for the z-axis. The voxel resolution is set to $0.2\,$m, resulting in 3D occupancy voxels of size $512\times 512\times 40$.
We compare our EfficientOCF against four SOTA methods: PowerBEV~\cite{li2023powerbev}, OpenOccupancy~\cite{wang2023openoccupancy}, OccFormer~\cite{zhang2023occformer}, and OCFNet~\cite{ma2024cam4docc}. 
All models are trained for $15$ epochs using AdamW optimizer~\cite{kingma2014adam} with an initial learning rate $3\,\times\,10^{-4}$ and a weight decay of $0.01$. Eight A100 GPUs are used for training with a batch size of $8$.

\subsection{Occupancy Forecasting Assessment}
\label{sec:assessment}
\vspace{-0.1cm}

We first assess the occupancy forecasting performance of EfficientOCF and baseline methods, as shown in Tab.~\ref{tab: comparison}. 
The nuScenes-Occupancy dataset provides fine-grained labels for the nuScenes dataset, allowing us to conduct evaluations based on fine-grained voxels. In contrast, the Lyft-Level5 dataset only contains bounding box annotations for movable objects, and thus the evaluation on this dataset is based on inflated objects where the $\text{C-IoU}$ metric is not available.
Notably, the vanilla PowerBEV only forecasts occupancy states in 2D space. Therefore, we lift its 2D results to 3D space by assigning a fixed height value to compute the 3D metrics. This fixed height for each test scene is determined by averaging the heights of all general movable objects.
For OpenOccupancy and OccFormer, we estimate the occupancy states of the present frame and then use it as the prediction of all future timesteps based on the static-world model~\cite{ma2024cam4docc}.
Besides, for OpenOccupancy, OccFormer, and OCFNet, which are end-to-end 3D occupancy perception methods, we first calculate 3D metrics and then compress the 3D results to BEV space to calculate the 2D metrics.
Following~\cite{ma2024cam4docc}, we additionally report the performance of OCFNet and our EfficientOCF when trained on limited $1/6$ of the training sequences, denoted as OCFNet$^{-}$ and EfficientOCF$^{-}$.
Tab.~\ref{tab: comparison} shows that our proposed EfficientOCF estimates the present occupancy states and forecasts future counterparts better than all the other baselines on the nuScenes, nuScenes-Occupancy, and Lyft-Level5 datasets in both 2D and 3D evaluation metrics. In particular, EfficientOCF improves the proposed C-IoU of the baselines more significantly than the other existing metrics. It is also notable that EfficientOCF$^{-}$ trained on limited data even outperforms the baseline OCFNet supervised by the holistic training set on all 3D fine-grained metrics. This validates that our method more effectively captures diverse motion patterns, even with only partial scenario perception, making it practically useful when training data is limited.

\begin{figure}
  \centering
  \captionsetup{aboveskip=1pt, belowskip=0pt}
  \includegraphics[width=0.9\linewidth]{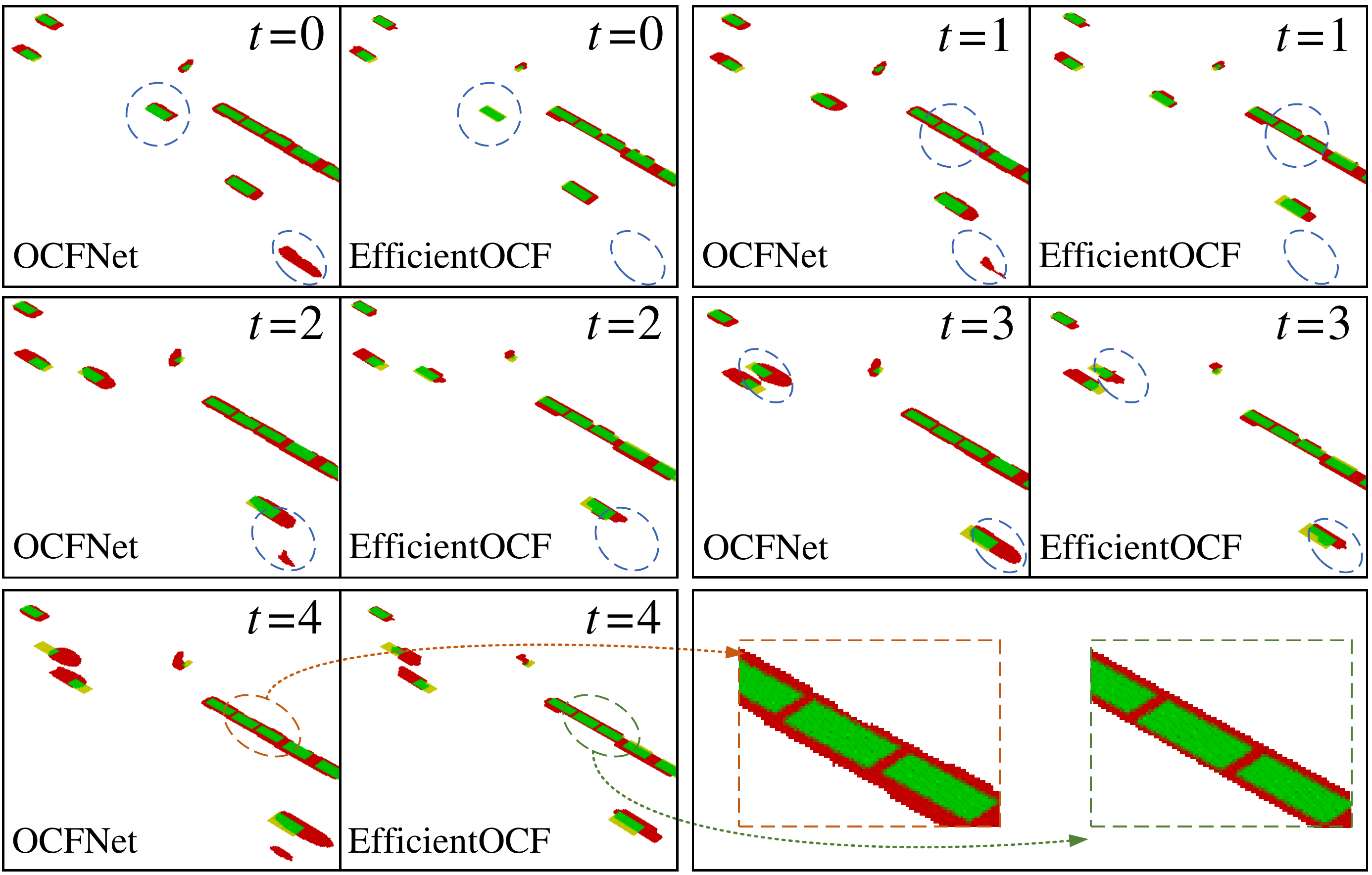}
  \caption{Visualization of TP (green), FP (red), and FN (yellow) results against GT at continuous timesteps.}
  \label{fig: tp}
  \vspace{-0.5cm}
\end{figure}

We further report the forecasting performance with different future time horizons. As shown in Tab.~\ref{tab: time}, our EfficientOCF remains the best performance within the future time horizons $\{\text{0.5\,s, 1.0\,s, 1.5\,s, 2.0\,s}\}$ on the nuScenes and nuScenes-Occupancy datasets, and $\{\text{0.2\,s, 0.4\,s, 0.6\,s, 0.8\,s}\}$ on the Lyft-Level5 dataset. Our temporal decoupling leads to more accurate OCF through instance forecasting associated with predicted flow at each timestep.
Fig.~\ref{fig: visualization} presents forecasting results of EfficientOCF at consecutive timesteps with colors from dark to light.
As can be seen, EfficientOCF captures better contours of movable objects (e.g., the first and the second samples of Fig.~\ref{fig: visualization}) compared to the SOTA baseline OCFNet which basically overlooks shape details. This is achieved by spatial decoupling, which provides more detailed 3D structures inherent in the combination of BEV occupancy and heights. 
Moreover, we also demonstrate that EfficientOCF can generate more TP and fewer FP predictions in 2D space by Fig.~\ref{fig: tp}, which also contributes to the higher forecasting accuracy of our EfficientOCF.
Besides, it can be seen from the nusc-occ ground-truth (GT) of Fig.~\ref{fig: visualization} that the publicly available fine-grained GT annotations are not always accurate due to LiDAR sparsity. This reflects the need for our new metrics C-IoU.
More visualization is in the supplementary.

\vspace{-0.1cm}
\subsection{Ablation Studies}
\label{sec:ablation}
\vspace{-0.1cm}

We conduct several ablation studies to validate the effectiveness of different components in EfficientOCF. All the 3D evaluation metrics are derived by comparing fine-grained predictions with corresponding ground-truth labels.

\textbf{Spatiotemporal decoupling.}~We refer the ablation study on spatial decoupling to Tab.~\ref{tab: comparison}, where our proposed EfficientOCF outperforms OCFNet that directly implements occupancy forecasting in dense 3D OCF representation. Here we further ablate temporal decoupling by removing the instance-aware refinement in EfficientOCF shown in Tab.~\ref{tab: temporal}. We conduct a baseline EfficientOCF$^\ddag$ that directly lifts the initial 2D OCF results with height values. As shown, our instance-aware temporal decoupling improves the initial OCF performance of the prediction module by absolute 1.64 and 2.55 percentage points in 3D $\tilde{\text{IoU}}_\text{f}$ and $\tilde{\text{C-IoU}}_\text{f}$ on the nuScenes-Occupancy dataset. This demonstrates that instance forecasting encompassing flow-based association provides the more accurate occupancy estimation for each decoupled timestep.

\textbf{Adaptive dual pooling.}~We compare our adaptive dual pooling (ADP) to different pooling strategies that compress 3D features to 2D space. Tab.~\ref{tab: transformation} shows that EfficientOCF with adaptive weighting for the combination of average and max pooling outperforms the baseline with only one single pooling strategy for 3D-2D transformation. The max pooling contributes more to forecasting accuracy than the average pooling, likely because height values and max-pooled features both capture prominent structural changes along the z-axis. 
We also deploy our ADP strategy as the 3D-2D transformation module in PowerBEV, and it enhances the performance of the vanilla PowerBEV in 2D $\text{IoU}_\text{c}$ and $\text{IoU}_\text{f}$ (2.0\,s) by absolute 5.16 and 5.42 percentage points. This validates that other OCF methods can also benefit from our proposed ADP strategy. 

\begin{table}[t]
\scriptsize
\setlength{\tabcolsep}{2pt}
\center
\captionsetup{aboveskip=2pt, belowskip=0pt}
\renewcommand\arraystretch{0.9}
\caption{Ablation on temporal decoupling}
\begin{tabular}{l|ccc|cccccc}
\toprule
\multicolumn{1}{l|}{\multirow{2}{*}{Approach}}   & \multicolumn{3}{c|}{2D} & \multicolumn{6}{c}{3D} \\ \cmidrule{2-10} 
\multicolumn{1}{c|}{}                                                                               & $\text{IoU}_\text{c}$    & $\text{IoU}_\text{f}$  & $\tilde{\text{IoU}}_\text{f}$  & $\text{IoU}_\text{c}$  & $\text{IoU}_\text{f}$    & $\tilde{\text{IoU}}_\text{f}$  & $\text{C-IoU}_\text{c}$  & $\text{C-IoU}_\text{f}$ & $\tilde{\text{C-IoU}}_\text{f}$  \\ \cmidrule{1-10}
EfficientOCF$^\ddag$   & 39.44   & 35.89   & 36.61   & 19.98   & 17.31   & 17.86    & 45.52   & 42.90   & 43.44   \\
EfficientOCF   & \textbf{39.93}  & \textbf{36.15}  & \textbf{36.93}  & \textbf{21.28}  & \textbf{19.02}  & \textbf{19.50}  & \textbf{47.53}  & \textbf{45.57}  & \textbf{45.99}\\   \bottomrule
\end{tabular}
\label{tab: temporal}
\vspace{-0.4cm}
\end{table}

\begin{table}[t]
\scriptsize
\setlength{\tabcolsep}{1.8pt}
\center
\captionsetup{aboveskip=2pt, belowskip=0pt}
\renewcommand\arraystretch{0.9}
\caption{Ablation on adaptive dual pooling}
\begin{tabular}{l|ccc|cccccc}
\toprule
\multicolumn{1}{l|}{\multirow{2}{*}{Approach}}   & \multicolumn{3}{c|}{2D} & \multicolumn{6}{c}{3D} \\ \cmidrule{2-10} 
\multicolumn{1}{c|}{}                                                                               & $\text{IoU}_\text{c}$    & $\text{IoU}_\text{f}$  & $\tilde{\text{IoU}}_\text{f}$  & $\text{IoU}_\text{c}$  & $\text{IoU}_\text{f}$    & $\tilde{\text{IoU}}_\text{f}$  & $\text{C-IoU}_\text{c}$  & $\text{C-IoU}_\text{f}$ & $\tilde{\text{C-IoU}}_\text{f}$  \\ \cmidrule{1-10}
Average Pooling   & 38.04   & 34.79   & 35.45   & 19.29   & 16.77   & 17.29    & 44.02   & 41.71   & 42.18   \\
Max Pooling   & 38.84   & 35.53   & 36.20   & 19.42   & 16.90   & 17.41    & 44.69   & 42.33   & 42.81   \\
ADP   & \textbf{39.93}  & \textbf{36.15}  & \textbf{36.93}  & \textbf{21.28}  & \textbf{19.02}  & \textbf{19.50}  & \textbf{47.53}  & \textbf{45.57}  & \textbf{45.99}\\   \bottomrule
\end{tabular}
\label{tab: transformation}
\vspace{-0.4cm}
\end{table}

\textbf{Multiple heads.}~We ablate the devised heads in the prediction module of EfficientOCF in Tab.~\ref{tab: head}. The inclusion of the height head allows the transformation of forecasting results from 2D to 3D space, and provides more structure-aware supervision signals. It thus significantly improves OCF performance on all the metrics. Furthermore, the introduction of the flow head also provides additional enhancements because flow prediction enables the model to be aware of the concrete motion pattern of each voxel grid. Besides, the flow head helps to refine the initial occupancy following our proposed temporal decoupling.
We provide an additional ablation on different flow formats in Sec.~\ref{sec: supp_add} of the supplementary material.

\begin{table}[t]
\scriptsize
\setlength{\tabcolsep}{1.4pt}
\center
\captionsetup{aboveskip=2pt, belowskip=0pt}
\renewcommand\arraystretch{0.9}
\caption{Ablation on segmentation/flow/height heads}
\begin{tabular}{lll|ccc|cccccc}
\toprule
\multirow{2}{*}{\begin{tabular}[l]{@{}l@{}}Seg.\\ Head\end{tabular}} & \multirow{2}{*}{\begin{tabular}[l]{@{}l@{}}Height\\ Head\end{tabular}} & \multirow{2}{*}{\begin{tabular}[l]{@{}l@{}}Flow\\ Head\end{tabular}} & \multicolumn{3}{c|}{2D} & \multicolumn{6}{c}{3D} \\ \cmidrule{4-12} 
    &    &    & $\text{IoU}_\text{c}$    & $\text{IoU}_\text{f}$  & $\tilde{\text{IoU}}_\text{f}$  & $\text{IoU}_\text{c}$  & $\text{IoU}_\text{f}$    & $\tilde{\text{IoU}}_\text{f}$  & $\text{C-IoU}_\text{c}$  & $\text{C-IoU}_\text{f}$   & $\tilde{\text{C-IoU}}_\text{f}$  \\ \cmidrule{1-12}
\checkmark    &    &    & 38.48   & 35.26   & 35.93   & -   & -    & -    & -    & -    & -    \\
\checkmark  & \checkmark  &    & 38.55    & 35.33    & 35.98   & 19.41   & 16.89   & 17.41  & 44.09   & 41.83   & 42.29 \\
\checkmark  & \checkmark  & \checkmark    & \textbf{39.93}  & \textbf{36.15}  & \textbf{36.93}  & \textbf{21.28}  & \textbf{19.02}  & \textbf{19.50}  & \textbf{47.53}  & \textbf{45.57}  & \textbf{45.99}\\ \bottomrule
\end{tabular}
\label{tab: head}
\vspace{-0.4cm}
\end{table}

\subsection{Instance Prediction Assessment}
\label{sec:flow}
\vspace{-0.1cm}

This experiment presents the 3D instance prediction performance of SOTA OCFNet and our EfficientOCF with the above-mentioned VPQ metric.
As shown in Tab.~\ref{tab: vpq}, EfficientOCF improves the VPQ of OCFNet forecasting by absolute 1.42 and 4.08 percentage points in $\text{VPQ}^\text{bb}$ and $\text{VPQ}^\text{fg}$ respectively on nuScenes, and by 8.40 percentage points on Lyft-Level5. EfficientOCF$^{-}$ still outperforms OCFNet$^{-}$ for both datasets.
EfficientOCF forecasts instances in 2D space which are further lifted to 3D counterparts following our proposed spatiotemporal decoupling. OCFNet directly implements instance association in dense 3D formats, leading to more prediction artifacts and uncertainties.

\begin{table}[t]
\scriptsize
\setlength{\tabcolsep}{11.5pt}
\center
\captionsetup{aboveskip=2pt, belowskip=0pt}
\renewcommand\arraystretch{0.9}
\caption{Evaluation on 3D instance prediction}
\begin{tabular}{l|cc|c}
\toprule
\multicolumn{1}{l|}{\multirow{2}{*}{Approach}} & \multicolumn{2}{c|}{nuScenes}  & \multicolumn{1}{c}{Lyft-Level5} \\ \cmidrule{2-4} 
\multicolumn{1}{c|}{}                                                                               & $\text{VPQ}^\text{bb}$   & $\text{VPQ}^\text{fg}$  & $\text{VPQ}^\text{bb}$ \\ \cmidrule{1-4}
OCFNet$^{-}$   & 13.68   & 4.25     & 24.11   \\
OCFNet   & 18.53   & 6.61     & 28.35   \\ \cmidrule{1-4}
EfficientOCF$^{-}$ (ours)   & 14.11   & 6.50     & 27.06   \\
EfficientOCF (ours)   & \textbf{19.95}   & \textbf{10.69}     & \textbf{36.75}   \\   \bottomrule
\end{tabular}
\label{tab: vpq}
\vspace{-0.3cm}
\end{table}

\subsection{Inference Time}
\label{sec:runtime}
\vspace{-0.1cm}

We compare the runtime of OCFNet and our EfficientOCF on 3D OCF shown in Tab.~\ref{tab: runtime}. 
As seen, the processing time of EfficientOCF is slightly higher than OCFNet due to the additional 3D-2D transformation, while EfficientOCF achieves a faster overall inference speed (82.33\,ms/12.15\,Hz) compared to OCFNet. 
Moreover, we find that EfficientOCF reduces memory consumption by 478\,MB compared to OCFNet. This is because EfficientOCF utilizes lightweight 2D encoders and decoders in the perception module to achieve 3D occupancy forecasting by lifting 2D results to 3D space using height values. In contrast, OCFNet uses heavy 3D encoders and decoders with dense 3D representation, leading to more inference time.

\begin{table}[t]
\scriptsize
\setlength{\tabcolsep}{6pt}
\center
\captionsetup{aboveskip=2pt, belowskip=0pt}
\renewcommand\arraystretch{0.9}
\caption{Comparison of inference time}
\begin{tabular}{l|ccc|c}
\toprule
\multicolumn{1}{l|}{\multirow{2}{*}{Approach}}   & \multicolumn{4}{c}{Inference time [ms]} \\ \cmidrule{2-5} 
\multicolumn{1}{c|}{}     & Perception  & Processing  & Prediction    & Total   \\ \cmidrule{1-5} 
OCFNet   & 107.05   & \textbf{4.18}   & 13.15   & 124.38   \\
EfficientOCF (ours)     & \textbf{64.78}    & 5.05   & \textbf{12.50}   & \textbf{82.33}   \\ \bottomrule
\end{tabular}
\label{tab: runtime}
\vspace{-0.4cm}
\end{table}


\section{Conclusion}

This paper proposes spatiotemporal decoupling for efficient vision-based occupancy forecasting (OCF). We achieve spatial decoupling by transforming 3D dense representations into compact 2D BEV occupancy with height values. We also propose using temporal decoupling to refine the end-to-end OCF by step-wise instance forecasting along the time axis.
Based on our proposed spatiotemporally decoupled representation, we further develop EfficientOCF, a lightweight network that concurrently forecasts future 2D occupancy states and height values of general movable objects, effectively bridging 2D predictions into the 3D space. It also refines initial OCF results with instance forecasting at each individual timestep.
Additionally, we introduce new evaluation metrics providing a comprehensive assessment of OCF performance for datasets with incomplete annotations.
Our experiments on extensive evaluation metrics demonstrate that EfficientOCF outperforms existing methods in both accuracy and efficiency, achieving SOTA results in both 2D and 3D spaces.
We hope that the spatiotemporal decoupling pipeline proposed in this work can be used as a foundation paradigm in the field of vision-based OCF.

\bibliographystyle{ieeetr}


\footnotesize{
\bibliography{new}}

\newpage
\setcounter{section}{0}
\setcounter{figure}{0}
\setcounter{table}{0}
\renewcommand{\thesection}{\arabic{section}}
\renewcommand{\thefigure}{\arabic{figure}}
\renewcommand{\thetable}{\arabic{table}}
\renewcommand{\thesection}{\arabic{section}}

\begin{flushleft}  
    \huge Supplementary Material
\end{flushleft}

\begin{table*}[t]
\scriptsize
\setlength{\tabcolsep}{1.4pt}
\center
\renewcommand\arraystretch{1}
\caption{Ablation on temporal decoupling in different time horizons}
\begin{tabular}{l|c|cccc|c|cccc|c|cccc|c|cccc}
\toprule
\multicolumn{1}{l|}{\multirow{3}{*}{Approach}}   & \multicolumn{10}{c|}{nuScenes} & \multicolumn{10}{c}{nuScenes \& nuScenes-Occupancy}\\ \cmidrule{2-21}  
\multicolumn{1}{c|}{}  & \multicolumn{1}{c|}{$\text{IoU}_\text{c}\,(\text{2D})$}   & \multicolumn{4}{c|}{$\text{IoU}_\text{f}\,(\text{2D})$} & \multicolumn{1}{c|}{$\text{IoU}_\text{c}\,(\text{3D})$}   & \multicolumn{4}{c|}{$\text{IoU}_\text{f}\,(\text{3D})$}   & \multicolumn{1}{c|}{$\text{IoU}_\text{c}\,(\text{3D})$} 
 & \multicolumn{4}{c|}{$\text{IoU}_\text{f}\,(\text{3D})$}  & \multicolumn{1}{c|}{$\text{C-IoU}_\text{c}\,(\text{3D})$}  & \multicolumn{4}{c}{$\text{C-IoU}_\text{f}\,(\text{3D})$} \\ \cmidrule{2-21} 
\multicolumn{1}{c|}{} &   & 0.5s  & 1.0s  & 1.5s  & 2.0s  &   & 0.5s  & 1.0s  & 1.5s  & 2.0s  &   & 0.5s  & 1.0s  & 1.5s  & 2.0s  &   & 0.5s  & 1.0s  & 1.5s  & 2.0s  \\ \cmidrule{1-21}
EfficientOCF$^\ddag$    & 39.44    & 38.40    & 37.52    & 36.71    & 35.89    & 32.19    & 31.63    & 30.82    & 30.45    & 30.08    & 19.98    & 18.61    & 18.21    & 17.79    & 17.31    & 45.52    & 45.12    & 44.35    & 43.65    & 42.90    \\
EfficientOCF    & \textbf{39.93}   & \textbf{38.80}  & \textbf{37.85}    & \textbf{36.98}    & \textbf{36.15}  & \textbf{35.60}  & \textbf{34.81}  & \textbf{34.07}  & \textbf{33.37}  & \textbf{32.73}  & \textbf{21.28}  & \textbf{20.21}  & \textbf{19.86}   & \textbf{19.46}    & \textbf{19.02}    & \textbf{47.53}   & \textbf{47.63}   & \textbf{46.91}   & \textbf{46.24}   & \textbf{45.57}      \\ \cmidrule{1-21}
Improvement   & 0.49    & 0.40    & 0.33    & 0.27    & 0.26    & 3.41    & 3.18    & 3.25    & 2.92    & 2.65    & 1.30    & 1.60    & 1.65    & 1.67    & 1.71    & 2.01    & 2.51    & 2.59    & 2.59    & 2.67   \\ \bottomrule
\end{tabular}
\label{tab: supp_temporal}
\vspace{-0.2cm} 
\end{table*}

\begin{table*}[t]
\scriptsize
\setlength{\tabcolsep}{1.4pt}
\center
\renewcommand\arraystretch{1}
\caption{Ablation on adaptive dual pooling and flow formats in different time horizons}
\begin{tabular}{l|c|cccc|c|cccc|c|cccc|c|cccc}
\toprule
\multicolumn{1}{l|}{\multirow{3}{*}{Approach}}   & \multicolumn{10}{c|}{nuScenes} & \multicolumn{10}{c}{nuScenes \& nuScenes-Occupancy}\\ \cmidrule{2-21}  
\multicolumn{1}{c|}{}  & \multicolumn{1}{c|}{$\text{IoU}_\text{c}\,(\text{2D})$}   & \multicolumn{4}{c|}{$\text{IoU}_\text{f}\,(\text{2D})$} & \multicolumn{1}{c|}{$\text{IoU}_\text{c}\,(\text{3D})$}   & \multicolumn{4}{c|}{$\text{IoU}_\text{f}\,(\text{3D})$}   & \multicolumn{1}{c|}{$\text{IoU}_\text{c}\,(\text{3D})$} 
 & \multicolumn{4}{c|}{$\text{IoU}_\text{f}\,(\text{3D})$}  & \multicolumn{1}{c|}{$\text{C-IoU}_\text{c}\,(\text{3D})$}  & \multicolumn{4}{c}{$\text{C-IoU}_\text{f}\,(\text{3D})$} \\ \cmidrule{2-21} 
\multicolumn{1}{c|}{} &   & 0.5s  & 1.0s  & 1.5s  & 2.0s  &   & 0.5s  & 1.0s  & 1.5s  & 2.0s  &   & 0.5s  & 1.0s  & 1.5s  & 2.0s  &   & 0.5s  & 1.0s  & 1.5s  & 2.0s  \\ \cmidrule{1-21}
Average Pooling    & 38.04    & 37.11    & 36.32    & 35.55    & 34.79    & 33.80    & 33.32   & 32.70   & 32.10   & 31.53    & 19.29    & 18.00    & 17.65    & 17.24    & 16.77    & 44.02    & 43.72    & 43.08    & 42.43    & 41.71    \\
Max Pooling    & 38.84    & 37.91    & 37.11    & 36.33    & 35.53    & 34.52    & 34.05    & 33.44    & 32.84    & 32.23    & 19.42    & 18.13    & 17.77    & 17.36    & 16.90    & 44.69    & 44.42    & 43.74    & 43.07    & 42.33    \\
ADP    & \textbf{39.93}   & \textbf{38.80}  & \textbf{37.85}    & \textbf{36.98}    & \textbf{36.15}  & \textbf{35.60}  & \textbf{34.81}  & \textbf{34.07}  & \textbf{33.37}  & \textbf{32.73}  & \textbf{21.28}  & \textbf{20.21}  & \textbf{19.86}   & \textbf{19.46}    & \textbf{19.02}    & \textbf{47.53}   & \textbf{47.63}   & \textbf{46.91}   & \textbf{46.24}   & \textbf{45.57}      \\ \cmidrule{1-21} 
\multicolumn{1}{c|}{}  & \multicolumn{1}{c|}{$\text{IoU}_\text{c}\,(\text{2D})$}   & \multicolumn{4}{c|}{$\text{IoU}_\text{f}\,(\text{2D})$} & \multicolumn{1}{c|}{$\text{IoU}_\text{c}\,(\text{3D})$}   & \multicolumn{4}{c|}{$\text{IoU}_\text{f}\,(\text{3D})$}   & \multicolumn{1}{c|}{$\text{IoU}_\text{c}\,(\text{3D})$} 
 & \multicolumn{4}{c|}{$\text{IoU}_\text{f}\,(\text{3D})$}  & \multicolumn{1}{c|}{$\text{C-IoU}_\text{c}\,(\text{3D})$}  & \multicolumn{4}{c}{$\text{C-IoU}_\text{f}\,(\text{3D})$} \\ \cmidrule{2-21} 
\multicolumn{1}{c|}{} &   & 0.5s  & 1.0s  & 1.5s  & 2.0s  &   & 0.5s  & 1.0s  & 1.5s  & 2.0s  &   & 0.5s  & 1.0s  & 1.5s  & 2.0s  &   & 0.5s  & 1.0s  & 1.5s  & 2.0s  \\ \cmidrule{1-21}
Backward    & 37.64    & 36.68    & 35.87    & 35.10    & 34.34    & 33.32    & 32.79    & 32.17    & 31.57    & 31.00    & 19.32    & 18.04    & 17.67    & 17.26    & 16.80    & 43.39    & 43.29    & 42.69    & 41.93    & 41.22    \\
Backward Centripetal    & \textbf{39.93}   & \textbf{38.80}  & \textbf{37.85}    & \textbf{36.98}    & \textbf{36.15}  & \textbf{35.60}  & \textbf{34.81}  & \textbf{34.07}  & \textbf{33.37}  & \textbf{32.73}  & \textbf{21.28}  & \textbf{20.21}  & \textbf{19.86}   & \textbf{19.46}    & \textbf{19.02}    & \textbf{47.53}   & \textbf{47.63}   & \textbf{46.91}   & \textbf{46.24}   & \textbf{45.57}      \\ \cmidrule{1-21}
Improvement   & 2.29    & 2.12    & 1.98    & 1.88    & 1.81    & 2.28    & 2.02    & 1.90    & 1.80    & 1.73    & 1.96    & 2.17    & 2.19    & 2.20    & 2.22    & 3.96    & 4.14    & 4.22    & 4.31    & 4.35   \\ \bottomrule
\end{tabular}
\label{tab: supp_ADP}
\vspace{-0.2cm} 
\end{table*}

\begin{table*}[t]
\scriptsize
\setlength{\tabcolsep}{1.4pt}
\center
\renewcommand\arraystretch{1}
\caption{Ablation on segmentation/flow/height heads in different time horizons}
\begin{tabular}{lll|c|cccc|c|cccc|c|cccc|c|cccc}
\toprule
\multirow{3}{*}{\begin{tabular}[l]{@{}l@{}}Seg.\\ Head\end{tabular}} & \multirow{3}{*}{\begin{tabular}[l]{@{}l@{}}Height\\ Head\end{tabular}} & \multirow{3}{*}{\begin{tabular}[l]{@{}l@{}}Flow\\ Head\end{tabular}} & \multicolumn{10}{c|}{nuScenes} & \multicolumn{10}{c}{nuScenes \& nuScenes-Occupancy} \\ \cmidrule{4-23} 
\multicolumn{3}{c|}{}  & \multicolumn{1}{c|}{$\text{IoU}_\text{c}\,(\text{2D})$}   & \multicolumn{4}{c|}{$\text{IoU}_\text{f}\,(\text{2D})$} & \multicolumn{1}{c|}{$\text{IoU}_\text{c}\,(\text{3D})$}   & \multicolumn{4}{c|}{$\text{IoU}_\text{f}\,(\text{3D})$}   & \multicolumn{1}{c|}{$\text{IoU}_\text{c}\,(\text{3D})$} 
 & \multicolumn{4}{c|}{$\text{IoU}_\text{f}\,(\text{3D})$}  & \multicolumn{1}{c|}{$\text{C-IoU}_\text{c}\,(\text{3D})$}  & \multicolumn{4}{c}{$\text{C-IoU}_\text{f}\,(\text{3D})$} \\ \cmidrule{4-23} 
\multicolumn{3}{c|}{}  &   & 0.5s  & 1.0s  & 1.5s  & 2.0s  &   & 0.5s  & 1.0s  & 1.5s  & 2.0s  &   & 0.5s  & 1.0s  & 1.5s  & 2.0s  &   & 0.5s  & 1.0s  & 1.5s  & 2.0s  \\ \cmidrule{1-23} 
\checkmark    &    &    & 38.48   & 37.63   & 36.92   & 35.84  & 35.26    & -    & -    & -    & -    & -   & -    & -    & -    & -    & -   & -    & -    & -    & -    & -    \\
\checkmark  & \checkmark  &    & 38.55    & 37.67    & 36.88   & 36.10    & 35.33  & 34.16    & 33.61   & 33.01    & 32.43   & 31.87   & 19.41   & 18.15   & 17.80   & 17.37   & 16.89  & 44.09   & 43.89   & 43.24   & 42.56   & 41.83 \\
\checkmark  & \checkmark  & \checkmark    & \textbf{39.93}   & \textbf{38.80}  & \textbf{37.85}    & \textbf{36.98}    & \textbf{36.15}  & \textbf{35.60}  & \textbf{34.81}  & \textbf{34.07}  & \textbf{33.37}  & \textbf{32.73}  & \textbf{21.28}  & \textbf{20.21}  & \textbf{19.86}   & \textbf{19.46}    & \textbf{19.02}    & \textbf{47.53}   & \textbf{47.63}   & \textbf{46.91}   & \textbf{46.24}   & \textbf{45.57}  \\ \bottomrule
\end{tabular}
\label{tab: supp_head}
\vspace{-0.4cm}
\end{table*}

\section{Additional Ablation Results}
\label{sec: supp_add}

In this section, we present additional ablation results for occupancy forecasting (OCF) mentioned in Sec.~\ref{sec:experiments} of the main text.
In Sec.~\ref{sec: supp_decoupling}, we ablate temporal decoupling for EfficientOCF. In Sec.~\ref{sec: supp_adp}, we then present the improvement of our proposed adaptive dual pooling (ADP) strategy for 3D-2D transformation in EfficientOCF. Next, in Sec.~\ref{sec: supp_flow}, we report performance changes with different flow formats. Finally, in Sec.~\ref{sec: supp_head}, we evaluate the contributions of multiple head combinations.

\subsection{Ablation on Temporal Decoupling}
\label{sec: supp_decoupling}

\begin{figure*}
\vspace{0.2cm}
  \centering
  \includegraphics[width=0.95\linewidth]{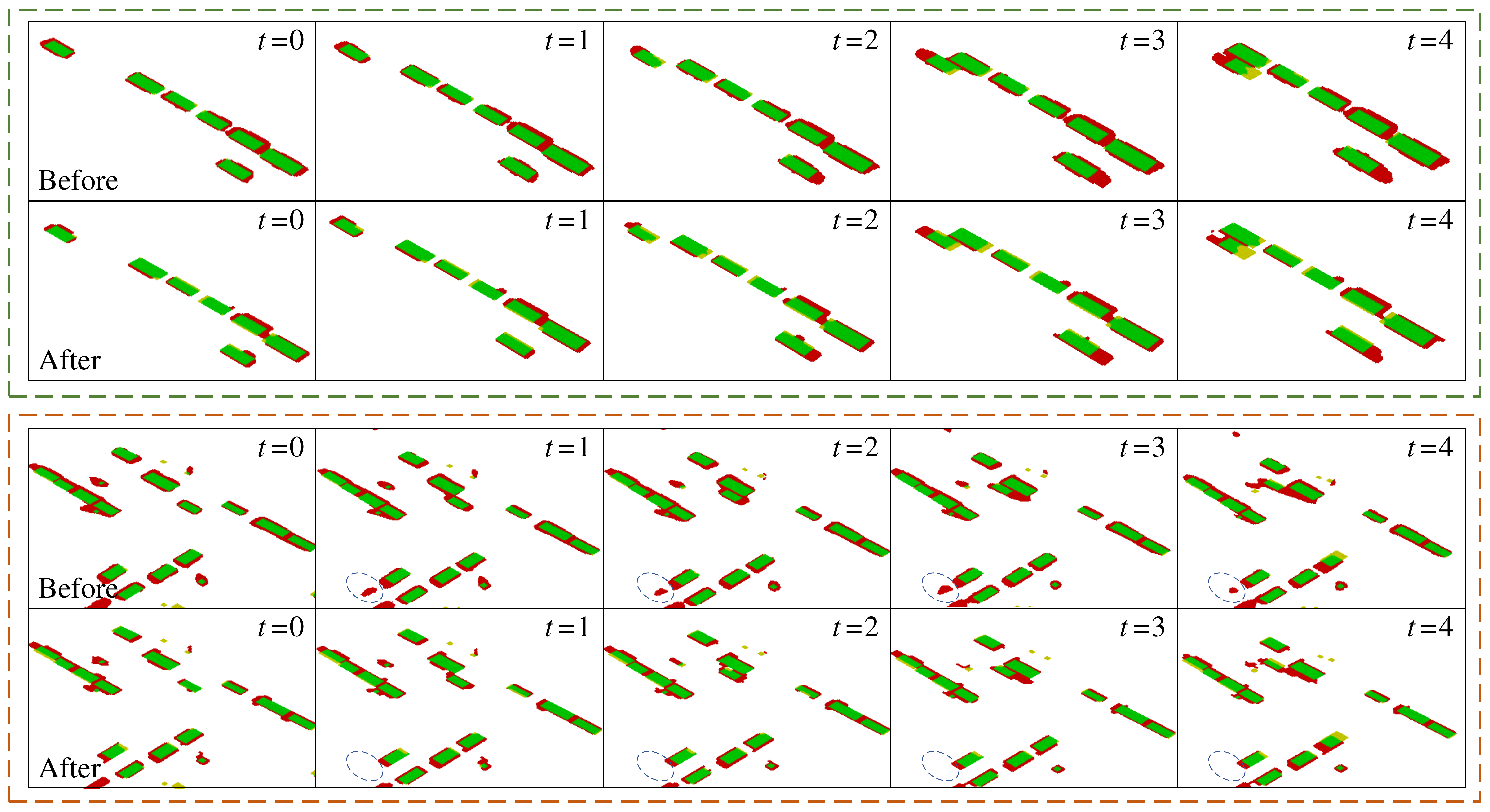}
  \caption{Visualization of TP (green), FP (red), and FN (yellow) results against ground truth at continuous timesteps before and after instance-aware refinement.}
  \label{fig: supp_tp}
  \vspace{-0.5cm}
\end{figure*}

We first study the proposed temporal decoupling by presenting supplementary OCF performance results of EfficientOCF over varying time horizons, as shown in Tab.~\ref{tab: supp_temporal}. The experimental results here are extensions of Tab.~\ref{tab: temporal} of the main text. As can be seen, our proposed temporal decoupling consistently enhances the baseline EfficientOCF$^\ddag$'s performance of all time horizons after using temporal refinement. 
Moreover, we observe that on the nuScenes-Occupancy dataset~\cite{wang2023openoccupancy}, the performance improvement becomes more significant with longer time horizons. In contrast, on the nuScenes dataset~\cite{caesar2020nuscenes}, the performance gains diminishment as the timestep increases.
This observation suggests that our proposed temporal decoupling enables EfficientOCF to focus better on the temporal dynamics of movable objects, thereby significantly enhancing its ability to forecast fine-grained occupancy states.
In Fig.~\ref{fig: supp_tp}, we further visualize the comparison of TP, FP, and FN results at continuous timesteps before and after instance-aware refinement. The two visualized cases show that our proposed temporal decoupling helps to increase TP and decrease FP predictions. Besides, it removes the predictions of non-existing movable objects, as shown in the blue circles in the lower case of Fig.~\ref{fig: supp_tp}.

\begin{figure*}
\vspace{0.2cm}
  \centering
  \includegraphics[width=0.95\linewidth]{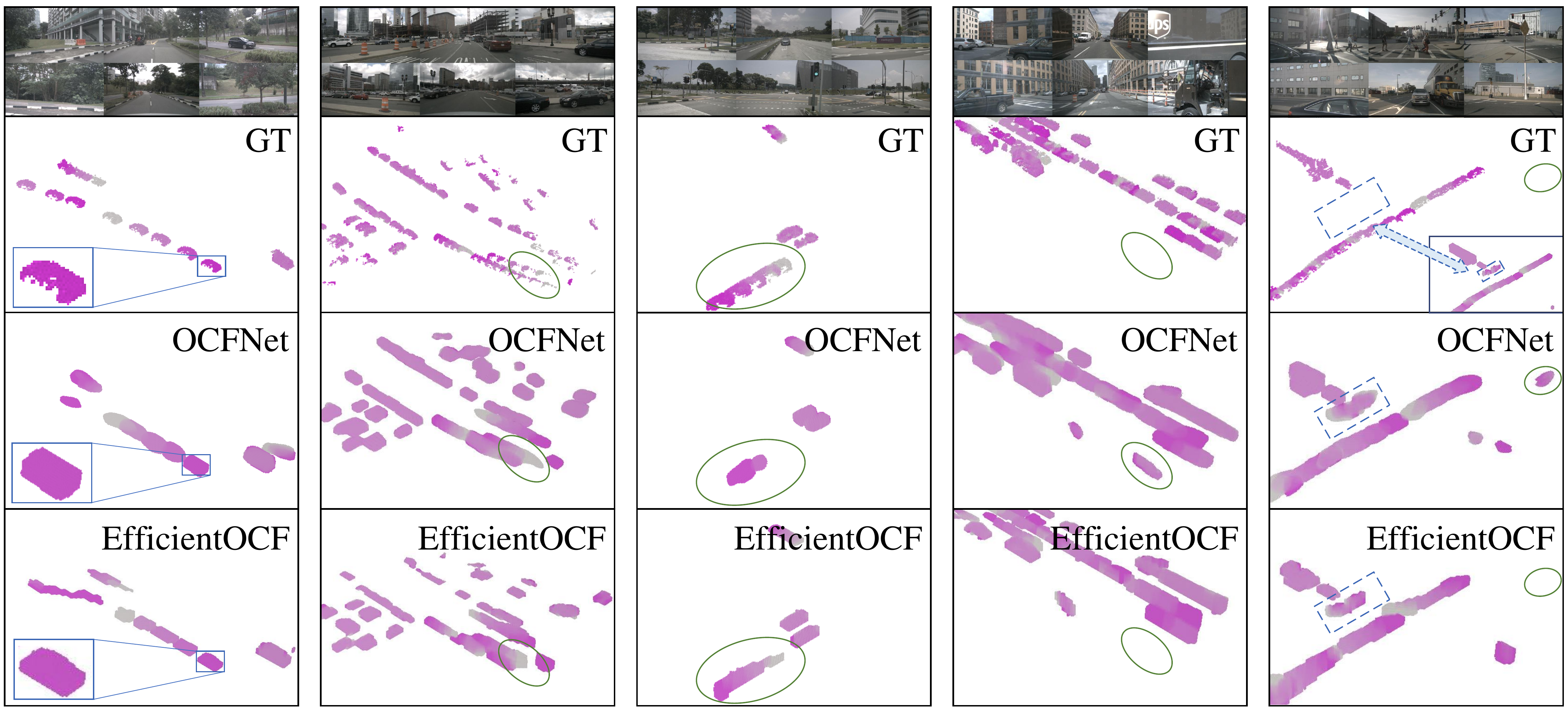}
  \caption{Visualization of OCF results of EfficientOCF, OCFNet, and ground truth from the nuScenes-Occupancy dataset \cite{wang2023openoccupancy}. The occupancy forecasting results and ground-truth labels from timesteps 0 to $N_\text{f}$ are assigned colors from dark to light.}
  \label{fig: supp_visualization}
  \vspace{-0.5cm}
\end{figure*}

\subsection{Ablation on Adaptive Dual Pooling}
\label{sec: supp_adp}

Here we provide an ablation study on adaptive dual pooling in different time horizons. This can be regarded as an extension of Tab.~\ref{tab: transformation} of the main text. As shown in Tab.~\ref{tab: supp_ADP}, the forecasting performance of EfficientOCF is comprehensively improved in all IoU metrics by ADP against the single pooling strategy, including the average pooling and max pooling. In particular, ADP improves $\text{C-IoU}_\text{c}$ prominently compared to other metrics.

\subsection{Ablation on Flow Formats}
\label{sec: supp_flow}

We further examine the impact of flow formats on OCF accuracy. The baseline model is constructed by substituting the backward centripetal flow in EfficientOCF with vanilla backward flow~\cite{liu2023multi, mahjourian2022occupancy}. As shown in Tab.\ref{tab: supp_ADP}, the backward centripetal flow yields superior occupancy forecasting performance, particularly in our proposed metrics, $\text{C-IoU}_\text{c}$ and $\text{C-IoU}_\text{f}$. These results indicate that backward centripetal flow is more robust to significant flow prediction errors, aligning with key findings in previous studies~\cite{li2023powerbev, ma2024cam4docc}.

\subsection{Ablation on Segmentation/Flow/Height Heads}
\label{sec: supp_head}

In Tab.~\ref{tab: supp_head}, we compare the OCF performance of EfficientOCF models with different combinations of prediction heads. The 3D OCF performance is not reported for the baseline with only the segmentation head, as it lacks height estimation to lift the predicted BEV occupancy into 3D space. The best performance is achieved when all heads are utilized across all time horizons. Notably, introducing the flow head results in a more significant performance improvement than the height head. This highlights that predicting future flow effectively captures the sequential motion of movable objects, thereby enhancing occupancy estimation at both present and future timesteps.

\section{Visualization of OCF Results}
\label{sec: supp_viz_ocf}

In this section, we present additional visualizations of fine-grained OCF results on the nuScenes-Occupancy dataset, comparing our proposed EfficientOCF with the SOTA baseline, OCFNet~\cite{ma2024cam4docc}. As shown in Fig.~\ref{fig: supp_visualization}, EfficientOCF consistently forecasts more accurate future occupancy states and captures more precise shapes of movable objects than OCFNet. In the first column of Fig.~\ref{fig: supp_visualization}, we highlight the occupancy of a moving vehicle within the blue box, where EfficientOCF predicts more detailed and accurate contours. The second and third columns demonstrate that EfficientOCF exhibits superior velocity awareness for movable objects compared to OCFNet. Additionally, the fourth and fifth columns reveal that OCFNet produces more false detections for invalid objects than our proposed EfficientOCF.
Notably, the fifth column also highlights missing annotations in nuScenes-Occupancy compared to the original nuScenes labels (bottom right of the ground-truth subfigure), likely due to LiDAR sparsity. This underscores the necessity of our new metrics, $\text{C-IoU}$, for evaluating fine-grained occupancy forecasting.

\end{document}